\UseRawInputEncoding

\documentclass{article}
\usepackage[T1]{fontenc}
\usepackage[utf8]{inputenc}
\usepackage{textcomp}
\usepackage{microtype}
\usepackage{amssymb}
\usepackage{amsmath}
\usepackage{graphicx}
\usepackage{subfigure}
\usepackage{array}
\usepackage{booktabs} %
\usepackage{longtable}
\usepackage{caption}
\DeclareMathOperator{\Ber}{Ber}

\usepackage{hyperref}
\newcommand{\R}{\mathbb{R}}
\usepackage[main, final]{neurips_2026}
\PassOptionsToPackage{numbers,compress}{natbib}

\newcommand{\one}{\mathbf{1}}
\usepackage{bm}
\usepackage{amsmath}
\usepackage{amssymb}
\usepackage{mathtools}
\usepackage{amsthm}
\usepackage{enumitem}
\usepackage{float}
\usepackage{comment}
\usepackage{xcolor}
\usepackage{tcolorbox}
\usepackage{listings}
\tcbuselibrary{listings,skins,breakable}
\lstdefinelanguage{Prompts}{morestring=[b]", sensitive=true}
\lstdefinestyle{promptstyle}{
  language=Prompts,
  basicstyle=\ttfamily\small,
  columns=fullflexible,
  breaklines=true,
  showstringspaces=false,
  keepspaces=true,
  upquote=true,
}
\newtcblisting{PromptBox}[2][]{%
  listing only,
  breakable,
  colback=black!1,
  colframe=black!50,
  fonttitle=\bfseries,
  title={#2},
  listing options={style=promptstyle},
  sharp corners,
  boxrule=0.6pt,
  left=1ex,right=1ex,top=1ex,bottom=1ex
}

\usepackage[capitalize,noabbrev]{cleveref}
\theoremstyle{plain}

\theoremstyle{definition}

\theoremstyle{remark}

\usepackage[disable,textsize=tiny]{todonotes}
\usepackage{fontawesome5}
\usepackage{tikz}

\title{Reasoning under uncertainty in humans \\
and language models}

\author{%
  \textbf{Gon\c{c}alo Guiomar}$^{1,2,*}$ \quad
  \textbf{Elia Torre}$^{1,2,*}$ \\[2pt]
  \textbf{Victoria Shavina}$^{1}$ \quad
  \textbf{Pehuen Moure}$^{1}$ \quad
  \textbf{Mario Giulianelli}$^{3}$ \quad
  \textbf{Shih-Chii Liu}$^{1}$ \quad
  \textbf{Valerio Mante}$^{1,2}$ \\[6pt]
  $^{1}$Institute of Neuroinformatics, University of Zurich and ETH Zurich, Switzerland \\
  $^{2}$ETH AI Center, ETH Zurich, Switzerland \\
  $^{3}$University College London, UK \\[4pt]
  $^{*}$Equal contribution.\\
  \texttt{\{goncalo.guiomar@ai.ethz.ch, etorre@ethz.ch\}}
}

\begin{document}
\addtocontents{toc}{\protect\setcounter{tocdepth}{-1}}

\maketitle

\begin{abstract}
Do language models make decisions under uncertainty like humans do? And if so, what role does \emph{extended reasoning} play in the underlying decision process? We answer this question by introducing an \emph{active probabilistic reasoning} task that cleanly separates \emph{sampling} (actively acquiring evidence) from \emph{inference} (integrating evidence towards a decision). Benchmarking humans and a broad set of contemporary LLMs against an optimal reference policy reveals a consistent pattern: extended reasoning is the key determinant of strong performance, driving large gains in inference, while yielding only modest improvements in active sampling. To explain these differences, we fit a behavioral model that captures systematic deviations from optimal Bayesian behavior through interpretable parameter families, placing humans and models in a shared low-dimensional \emph{cognitive space}. The resulting fits show how \emph{reasoning} shifts models toward human-like regimes of evidence accumulation and belief-to-choice mapping, and yield testable predictions about the latent dynamics that might drive each decision. Probing residual-stream activations of an open-weight reasoning model, we find that the geometry of internal representations tracks these predicted dynamics, linking behavior to representational correlates of the fitted latent dynamics.
\end{abstract}
\definecolor{trtred}{HTML}{CF1E4C}
\begin{center}
\vspace{-0.9em}
{\hypersetup{pdfborder={0 0 0}}%
\href{https://ai.trt-bench.org}{\raisebox{-0.5em}{\includegraphics[height=1.75em]{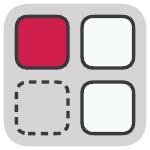}}\hspace{0.45em}\textcolor{trtred}{\texttt{ai.trt-bench.org}}}}
\vspace{-0.6em}
\end{center}

\section{Introduction}
Post-training LLMs to produce long intermediate \textit{reasoning traces} before committing to an answer can substantially improve performance on challenging benchmarks, often reaching near-human accuracy \citep{nye2021show, reynolds2021prompt, wei2022chain, kojima2022large, wang2022self, chen2025towards, jaech2024openai, guo2025deepseek, el2025competitive}. However, it remains unclear \emph{why} extended reasoning helps and \emph{how} these traces relate to the computations that actually produce the answer. Indeed, the traces can be unreliable and can diverge from the model's internal processes, implying that they do not necessarily describe the true algorithm a model applied in solving a task \citep{turpin2023language, lanham2023measuring, barez2025chain, chen2025reasoning, arcuschin2025chain}.

\begin{figure*}[!h]
\centering
\includegraphics[width=\linewidth]{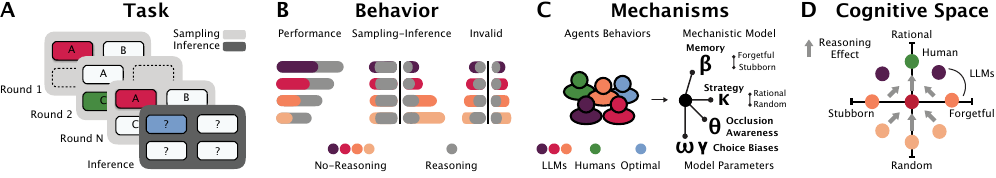}
\caption{\textbf{From task performance to latent cognitive variables.}
\textbf{A: Task.} An \emph{active probabilistic reasoning} task in which agents sequentially sample from up to four buttons (A–D), each revealing a binary outcome (RED/GREEN). One button is biased toward RED, while the others are unbiased. During $T$ \emph{sampling} rounds, agents sample evidence among the buttons available. During \emph{inference} round, agents indicate which button they believe is biased. An equivalent text-based version is used for LLMs (App.~\textbf{~\ref*{sec:prompts}}).
\textbf{B: Behavior.} Comparison of human and LLM behavior by \textit{overall performance}, \textit{sampling/inference quality}, and \textit{invalid choices}. These metrics reveal a broad spectrum of performance, with \emph{extended reasoning} improving overall success via enhanced \textit{inference}, while gains in \textit{sampling} remain limited.
\textbf{C: Mechanisms.} Behavioral model that captures deviations from optimal Bayesian inference using interpretable parameters: \textbf{Memory} ($\beta$), \textbf{Strategy} ($m,\kappa_f$), \textbf{Choice Bias} ($\omega,\gamma$), and \textbf{Occlusion Awareness} ($\theta_s,\theta_f$).
\textbf{D: Cognitive space.} These parameters define a shared low-dimensional space for humans and LLMs. \textit{Reasoning} shifts LLMs toward human-like \textit{inference} strategies, and tightens the gap in \textit{sampling} strategies.}
\label{fig:task}
\end{figure*}

More generally, algorithmic-level descriptions of the representations underlying LLM behavior are arguably missing for most current benchmarks \citep{moskvichev2023conceptarc}. Yet, a long line of work in cognitive neuroscience suggests that such descriptions could be critically important for any attempts to link a behavior to specific parameters and activations in the underlying circuits \citep{marr2010vision, lengyel2024marr}. Making explicit the algorithms and representations that lead to a particular behavior can yield \emph{interpretable latent variables} that are often more directly tied to activations than the observable behavior, and thus help bridge the gap between behavior and high-dimensional activations \citep{lengyel2024marr, ku2025using}. 

Such algorithmic descriptions are also needed to evaluate human–LLM alignment and how extended reasoning shapes it. Fine-tuning on human feedback and behavioral datasets can make LLMs reproduce human-like choices and signatures of classic cognitive models \citep{lake2017building, peterson2021using, wei2021finetuned, bai2022training, ouyang2022training, BinzSchulz2023-LLMasCognitiveModels, momente-etal-2025-triangulating, BinzEtAl2025-Centaur-Nature}. Yet, performance and behavioral similarity alone may not be sufficient to conclude that LLMs rely on the same latent computations and decision strategies as humans.

In this work, we take inspiration from cognitive science by developing a behavioral task that isolates two core elements of decision-making under uncertainty: \emph{sampling} and \emph{inference}. Despite its simple structure, our task is demanding, making it a sensitive benchmark for both humans and frontier LLMs. Behavior in our task can be explained in terms of the dynamics of a few, interpretable latent variables, which capture the agents' strategies and how they are shaped by \emph{reasoning}. Concretely, we make the following contributions:

\begin{itemize}[leftmargin=*,labelsep=0.5em,topsep=2pt,itemsep=0pt,parsep=0pt,partopsep=0pt]
    \item We introduce an \emph{active probabilistic reasoning task} that disentangles \emph{sampling} from \emph{inference} and is well-suited for algorithmic analysis of behavior (Section\textbf{~\ref{sec:task}}, Fig.\textbf{~\ref{fig:task}A}).
    \item We benchmark humans and a broad set of LLMs against an exact optimal reference policy, and find that top \textit{reasoning} models match or exceed humans in \emph{inference quality} but consistently under-perform in \emph{sampling quality} (Sections\textbf{~\ref{sec:bench}}--\textbf{\ref{sec:comparison}}, Fig.\textbf{~\ref{fig:task}B}; Fig.\textbf{~\ref{fig:bench}}).
    \item We develop a behavioral model that places humans and LLMs in a shared \emph{cognitive space} of interpretable parameters (\textbf{Memory}, \textbf{Strategy}, \textbf{Choice Bias}, \textbf{Occlusion Awareness}), shows how \emph{extended reasoning} shifts LLMs toward human-like strategies, and produces latent dynamics that yield testable predictions about internal computation (Sections~\textbf{\ref{sec:mech_model}},\textbf{\ref{sec:model_results}}; Fig.~\textbf{\ref{fig:task}C--D}; Fig.~\textbf{\ref{fig:mech_model}}).
    \item We relate behavior to residual-stream activations in \textit{gpt oss 20b}, which encode round-by-round trajectories tracking the latent dynamics predicted by the fitted model (Section~\textbf{\ref{sec:latent}}; Fig.~\textbf{\ref{fig:cognitive_space}}).
\end{itemize}

\section{Related work}
LLMs are increasingly studied as generalist decision-making agents that act in dynamic environments and adapt via \emph{in-context learning} (ICL) and \emph{in-context reinforcement learning} (ICRL) \citep{brown2020language, yao2022react, laskin2022ad, shinn2023reflexion, liu2023reason, liu2023agentbench, dong2024survey, moeini2025survey}. A complementary line treats LLMs as candidate models of human cognition, where behavior can potentially be linked to internal representations \citep{schrimpf2021neural, frank2023openly, binz2023turning, aher2023using, zhao2025large, loo2026llms}. These lines intersect in sequential decision-making, e.g. multi-armed bandits (MABs) \citep{slivkins2019introduction, lattimore2020bandit}, where LLMs can leverage feedback to improve choices in-context \citep{lee2023supervised, codaforno2023meta, krishnamurthy2024can, monea2025icrl, schubert2024context}. Moreover, alignment and fine-tuning on human feedback and behavioral datasets can substantially increase human-likeness, in some cases matching the decision dynamics of traditional cognitive models \citep{bai2022training, ouyang2022training, binz2024centaurfoundationmodelhuman, BinzEtAl2025-Centaur-Nature, momente-etal-2025-triangulating, SuHoGureckis2025IBT}. However, most evaluations emphasize the exploration–exploitation tradeoff or overall performance, offering limited leverage to distinguish strategies that produce similar behavior \citep{park2024regret, rahn2024east, nie2024evolve, felicioni2024uncertainty}. Our task departs from classical stochastic MABs by design: instead of conflating information gathering with reward maximization, it separates sampling from inference, which we find to be key to precisely assess alignment between LLMs and humans.

\section{Active probabilistic reasoning task}
\label{sec:task}
We introduce an \emph{active probabilistic reasoning} task (Fig\textbf{.~\ref{fig:task}A}) inspired by classical decision-making and $K$-armed bandit paradigms \citep{LaiRobbins1985Bandits, DawEtAl2006NN, NajemnikGeisler2005Nature, slivkins2019introduction, lattimore2020bandit}, but explicitly designed to disentangle \emph{sampling} from \emph{inference}. The task is well-suited for modeling at the algorithmic level and is simple compared to many current benchmarks \citep{jimenez2023swe, sprague2023musr, liu2023agentbench, zhou2023webarena, mialon2023gaia, phan2025humanity}, yet \emph{extended reasoning} is required for LLMs to perform well, and even leading models under-perform humans in key aspects of the task.

Each trial is an independent game with a variable number of \textit{sampling} rounds $T\in \{2,\dots,15\}$ followed by a single \textit{inference} round in which the agent must commit to a final decision. On each sampling round, the agent selects one of four buttons (A--D) and observes a binary outcome (RED/GREEN); at the start of game $g$, exactly one button is biased towards RED, emitting RED with probability $\alpha_B=0.9$, while the remaining buttons are unbiased with $\alpha_U=0.5$. The agent’s objective is to identify the biased button from the sampled evidence and report it in the inference round. Formally, the biased button is a latent variable $z_g\in\{1,\dots,4\}$; at round $t$ the agent chooses $a_{t}\in\{1,\dots,4\}$ and observes $x_{t}\in\{0,1\}$ (with $x=1$ for RED). To probe how agents adapt their sampling strategy to changing action sets, between $0$ and $3$ buttons are temporarily occluded on each round. These occlusions help distinguish posterior-guided sampling from choice persistence by requiring agents to select among alternative options when their preferred button becomes unavailable.

Unlike in standard bandits, sampling choices carry no reward: they only provide evidence about the latent hypothesis (\textit{which button is biased}), and performance depends solely on the final inference choice. This separation enables independent assessment of \textit{sampling quality} and \textit{inference quality}.

\section{Data collection} 
\label{sec:bench}
We collect behavioral data from both human participants and language models. Human participants played an interactive, graphical version of the task\footnote{\url{https://ai.trt-bench.org}}. For language models, we designed an equivalent \emph{text-based} version of the task, in which LLMs interact via token-defined choices under the same instructions provided to the human players (see App.~\textbf{\ref{sec:prompts}}).

\textbf{Humans.} We recruited $50$ human participants for a 1-hour live in-person competition. All participants provided written informed consent prior to the task and received instructions matching those used to prompt the LLMs (App.~\textbf{\ref{sec:prompts}}). Participants played multiple independent games with trial lengths sampled uniformly between $2$ and $15$ rounds. Of the $50$ participants, $46$ completed the full protocol of $100$ games each, yielding $4{,}600$ games in total (see App.~\textbf{\ref{appdx:humans}} for individual participant performance profiles).\looseness=-1

\begin{figure*}[!t]
    \centering
    \includegraphics[width=\textwidth]{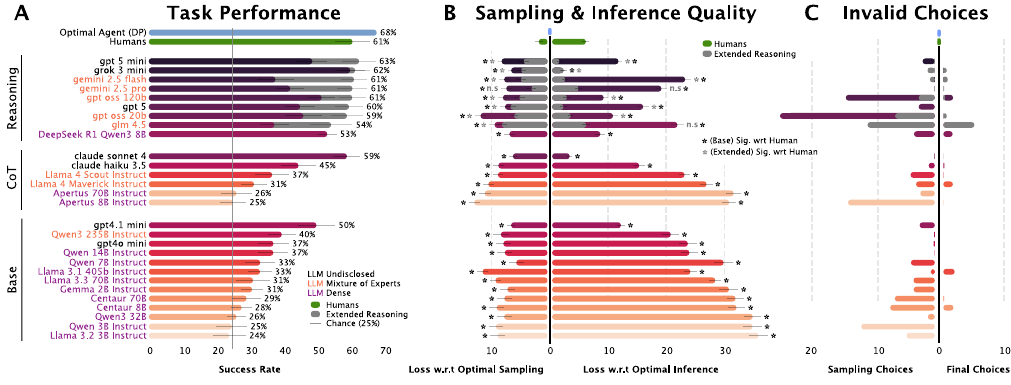}
    \caption{\textbf{Extended reasoning enables LLMs to match or exceed humans in inference, but not in sampling.} Models are grouped into \emph{Reasoning}, \emph{CoT}, and \emph{Base} (App.~\ref{app:llm_list}); model-name colors indicate undisclosed (black), mixture-of-experts (orange), or dense (purple) architectures. \textbf{A: Task performance.} Mean success rate across trial lengths $T\in\{2,\dots,15\}$. Humans are shown in green, the optimal dynamic-programming agent in blue, and extended-reasoning conditions as gray overlays. The vertical line marks chance (25\%). \textbf{B: Sampling and inference quality.} Performance losses relative to the optimal agent (lower is better). Sampling loss compares the optimal reference with MAP inference applied to each agent's sampled evidence; inference loss compares this counterfactual MAP performance with actual success. Error bars show bootstrap standard deviations ($B=1{,}000$). Asterisks indicate significance of differences from human performance (black: baseline; gray: extended reasoning; n.s.: not significant). \textbf{C: Invalid choices.} Fraction of invalid sampling and final choices, including occluded, malformed, or missing responses. Extended reasoning primarily improves inference, with smaller gains in sampling and fewer invalid choices.}
    \label{fig:bench}
\end{figure*}

\textbf{Large language models.} We evaluate a broad set of LLMs spanning state-of-the-art closed-source and open-weight models, covering \emph{dense} and \emph{Mixture-of-Experts} architectures across pre-trained, instruction-tuned and reasoning paradigms \citep{vaswani2017attention, schulman2017proximal, shoeybi2019megatron, wei2021finetuned, ouyang2022training, wei2022chain, shu2023exploitability, shao2024deepseekmath, cai2025survey}. Model details are in Figures~\ref{fig:bench}, \ref{fig:mech_model} and App.~\ref{app:llm_list}. We group models into three regimes: \emph{Reasoning} models with a controllable effort parameter, evaluated at \emph{low} and \emph{high} effort (\emph{extended reasoning}, gray bars; Fig.~\ref{fig:bench}A); \emph{CoT} models emitting a \textit{chain-of-thought} trace (max 500 tokens); and \emph{Base} models constrained to a single-token response (see App.~\ref{app:llm_list} and App.~\ref{app:sampling_sign}--\ref{app:inference_sign} for breakdowns). Each LLM and condition is evaluated on at least $1{,}400$ games, uniformly spanning $2$–$15$ rounds; model fits use a balanced subset of $100$ games per trial length ($1{,}400$ per condition).\looseness-1

\section{Optimal agent} 
\label{sec:optimal}
To evaluate and compare the performance of LLMs and humans, we introduce an optimal reference agent that maximizes expected task performance. Establishing the optimal strategy for \textit{inference} is straightforward: an agent should select the button with the highest posterior probability of being biased, following a \emph{Maximum-A-Posteriori} (MAP) decision rule \citep{griffiths2008bayesian}. A single game $g$ of $T$ rounds consists of a sequence of choice-evidence pairs $(a_{1:T},x_{1:T})$ together with a latent context variable $k\in\{1,\dots,K\}$. We represent beliefs by the posterior vector $\bm p_{t}\in \Delta^K$. We also define the likelihoods over the latent variable $k$ and emission rates $\{\alpha_B, \alpha_U\}$ by $\bm L_{t}\in \Delta^K$. 

By assuming a uniform prior, we can write the recursive normative posterior via Bayes' rule as (see App.~\ref{appdx:model_formulation}):\looseness=-1 
\begin{equation}
\bm p_{t}=\frac{\bm p_{t-1}\odot \bm L_{t}}{\mathbf{1}^\top(\bm p_{t-1}\odot \bm L_{t})},
\qquad
(\bm p_{0})_k=\tfrac{1}{4}.
\label{eq:normative_update}
\end{equation}
where $\odot$ denotes the Hadamard product. The MAP \emph{inference} policy at round $t$ selects the hypothesis with maximum posterior and is given by: 
\begin{equation}
a^{\mathrm{MAP}}_{t}
:= \arg\!\max_{k\in\{1,\dots,4\}} (\bm p_{t})_k .
\label{eq:map_policy}
\end{equation}

For our task's parameters, the optimal \emph{sampling} strategy is also MAP: sample the available cue with the highest posterior probability. Exact dynamic programming over posterior belief states, under the task's random occlusions, shows that MAP sampling followed by a MAP final choice attains the optimum for every game length (mean success $67.8\%$; App.~\textbf{\ref{appdx:dp_ceiling}}). This optimal agent is the reference for benchmarking and for decomposing performance into \textit{sampling} and \textit{inference} losses (Fig.~\textbf{\ref{fig:bench}}).

\section{Comparing human \& LLM behavior}
\label{sec:comparison}
To compare human and LLM performance, we employ four complementary metrics that capture distinct aspects of behavior: (1) overall task performance, (2) sampling quality, (3) inference quality, and (4) invalid choice rates.

\textbf{Overall task performance.} In Figure \textbf{\ref{fig:bench}A}, we quantify \emph{overall task performance} as the average \emph{success rate} across all trial lengths ($T \in \{2,\dots,15\}$), i.e., the fraction of games in which the agent correctly identifies the biased button at the inference round. Our task induces a broad distribution of performance across both humans and models with clear gains from \emph{extended reasoning} (gray vs.\ colored bars). On average, human participants achieve $\simeq 61\%$ success, comparable to the best-performing LLMs (\emph{gpt 5 mini}), with substantial inter-participant variability (App.~\textbf{\ref{appdx:humans}}). Task performance also varies substantially with trial length (App.~\textbf{\ref{appdx:score_evolution}}). Unlike most \textit{Base} configurations, all eight \textit{extended-reasoning} configurations exhibit a predicted improvement exceeding $5$~pp between $T=2$ and $T=15$ (mean gain $40.0$~pp). These gains exceed those of \textit{CoT} and \textit{Base} configurations by $29.4$~pp ($p_{\mathrm{Holm}}=0.0015$) and $37.1$~pp ($p_{\mathrm{Holm}}<0.0001$), respectively, suggesting that \textit{extended reasoning} improves the ability to leverage additional evidence in longer games.

\textbf{Sampling \& inference quality.} In Figure \textbf{\ref{fig:bench}B}, we quantify \emph{inference quality} as the performance gap between each agent and a counterfactual agent that keeps the agent's samples but performs MAP inference (right), and \emph{sampling quality} as the gap between this counterfactual and the optimal agent, which share the same MAP inference rule (left; lower is better for both). We find that \textit{inference quality} varies broadly across LLMs and is strongly correlated (negatively, because the metric is a loss) with overall performance ($r=-0.99$), whereas \textit{sampling quality} is more homogeneous across LLMs and a weaker predictor of performance ($r=-0.70$; both across all 36 LLM conditions). Performance gains from \textit{extended reasoning} are driven mainly by improved inference: across the 8 models tested under both reasoning conditions, reasoning reduced inference loss by $10.9$~pp on average versus only $3.0$~pp for sampling loss (95\% bootstrap CI: $[+3.5,+12.4]$~pp; paired permutation $p=0.015$). This asymmetry persists after excluding games containing invalid sampling choices: \textit{extended reasoning} reduces inference loss by $10.36$~pp compared with $3.87$~pp for sampling loss (paired difference: $6.50$~pp; exact one-sided paired sign-flip $p=0.043$). Several \textit{extended-reasoning} models match or exceed pooled human inference quality, whereas every LLM has a higher sampling loss than humans, significantly so for all but one condition (Fig.~\textbf{\ref{fig:bench}B}; see App.~\textbf{\ref{app:sampling_sign}},~\textbf{\ref{app:inference_sign}} for per-model statistics). This dissociation is not readily explained by token count alone: under \emph{high-reasoning}, sampling rounds consume more tokens than inference decisions ($\sim$8k--12k vs.\ $\sim$2k--5k for \emph{gpt oss 20b}; App.~\textbf{\ref{app:reasoning_analysis}}) yet benefit less, and representative traces suggest that inference gains involve explicit evidence integration (App.~\textbf{\ref{app:reasoning_traces}}). Humans also follow the optimal MAP sampling rule more closely than \textit{extended-reasoning} LLMs, especially when the most probable cue is occluded (App.~\textbf{\ref{app:posterior_sampling}}).

\textbf{Invalid choice rates.} Finally, in Figure \textbf{\ref{fig:bench}C}, we quantify \emph{invalid choices} as the proportion of \textit{sampling} or \textit{inference} choices that select an occluded button, fall outside the valid vocabulary (A–D), or are missing. Invalid responses are more frequent during sampling than inference and decrease under extended reasoning (Fig.~\textbf{\ref{fig:bench}C}). Because the human interface enforces valid choices, these rates measure LLM task compliance rather than a human--LLM difference.

\section{A behavioral model of latent cognitive variables}
\label{sec:mech_model}
\begin{figure*}[!t]
\centering
\includegraphics[width=\linewidth]{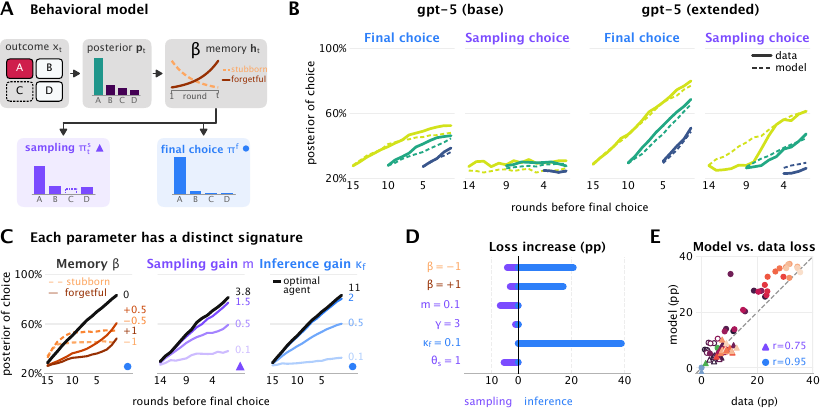}
\caption{\textbf{The behavioral model reproduces behavior, and each parameter has a distinct signature.}
\textbf{A: Behavioral model.} The posterior $\bm p_t$ is integrated into memory $\bm h_t$ with a kernel set by $\beta$ (light orange, dashed: \emph{stubborn}; dark orange: \emph{forgetful}), which drives the sampling ($\bm\pi^s_t$) and final ($\bm\pi^f$) policies.
\textbf{B: Posterior dynamics.} Posterior of the chosen cue in real (solid) and model-simulated (dashed) games of 15, 10 and 5 rounds.
\textbf{C: Parameter signatures.} Single-parameter perturbations of the model fitted to the optimal agent (black; 15-round games).
\textbf{D: Loss increase} in sampling (left) and inference (right) for these perturbations.
\textbf{E: Model vs.\ data loss.} Sampling ($\blacktriangle$) and inference ($\bullet$) loss of all 38 agents (open: \textit{extended reasoning}).}
\label{fig:model_summary}
\end{figure*}

The behavioral metrics above reveal systematic differences across LLMs, but not the processes that generate them. We therefore fit a compact behavioral model (Fig.~\textbf{\ref{fig:model_summary}A}) to test how reasoning changes \textit{sampling} and \textit{inference} strategies. The model is motivated by Bayesian ideal-observer and probabilistic cognition frameworks \citep{knill2004bayesian,sanborn2010rational}, and by views of \emph{in-context learning} as implicit inference over latent context variables \citep{xie2022explanation,panwar2024bayesianprism,zhang2023what}. On this normative backbone, it allows systematic deviations in \textbf{memory}, \textbf{strategy}, \textbf{choice bias/inertia}, and \textbf{occlusion awareness}. These explanations remain behavioral, but provide structured hypotheses about possible mechanisms. Below, we use the fitted parameters to compare agents and to guide the representational analyses of Sec.~\textbf{\ref{sec:latent}}.

\paragraph{Posterior of the chosen cue.} Because the Bayesian posterior $\bm p_t$ is known at every round, each choice can be summarized by the posterior of the chosen cue (Fig.~\textbf{\ref{fig:model_summary}B}). An optimal agent always chooses the most probable cue, so this readout rises across rounds for both its sampling and final choices. Without extended reasoning, \emph{gpt 5} samples almost independently of the evidence (the sampled cue stays near the $25\%$ chance level) and its final choice reaches only $52\%$ after $15$ rounds; with extended reasoning both rise, to $61\%$ for sampling and $80\%$ for the final choice. The parameters below shape these two readouts in distinct ways (Fig.~\textbf{\ref{fig:model_summary}C--D}).

\paragraph{Memory ($\beta$).}
As in sequential sampling models \citep{busemeyer1993decision,usher2001time}, choices depend on a non-optimal \emph{memory} $\bm h_t\in\mathbb R^K$ \citep{gold2007neural,hogarth1992order,MaKordingGoldreich2023BayesianModels}. Let $\bm p_t$ be the Bayesian posterior after round $t$, set $\bm h_0=\bm 0$, and define centered log-posterior increments
\begin{equation}
\Delta \bm h_t
:=
C(\log\bm p_t)-C(\log\bm p_{t-1}),
\qquad
C(\bm x)=\bm x-\langle \bm x\rangle\mathbf 1,
\end{equation}
with which we can define our memory integration process by applying a gain schedule to previous increments:\looseness=-1
\begin{equation}
\bm h_t
=
\sum_{u=1}^{t}w_{t,u}(\beta)\Delta \bm h_u,
\qquad
w_{t,u}(\beta)=
\begin{cases}
\exp[-\beta(t-u)], & \beta>0,\\
1, & \beta=0,\\
\exp[\beta(u-1)], & \beta<0.
\end{cases}
\label{eq:h_gamma_kernel}
\end{equation}
Thus $\beta=0$ recovers perfect accumulation, $\beta>0$ produces \emph{forgetful} recency weighting, and $\beta<0$ produces \emph{stubborn} primacy weighting where earlier evidence dominates; stubborn memory makes the final-choice posterior plateau early, forgetful memory makes it rise late (Fig.~\textbf{\ref{fig:model_summary}C}, left).

\paragraph{Strategy ($\kappa_s,\kappa_f$).}
We map beliefs to actions with a softmax/logit choice rule, following probabilistic-choice and bounded-rational decision models \citep{luce1959individual,mcfadden1974conditional,ortega2013thermodynamics} using time-varying \textit{sampling} gain $\kappa_s(t)$ and a static \textit{inference} gain parameter. For \textit{sampling}, as the outcome at round $t$ has not yet been observed, sampling depends on the pre-choice memory $\bm h_{t-1}$. Let $o_u\in\{0,1\}$ indicate whether round $u$ yielded a valid observation, and define
\begin{equation}
n_t
:=
1+\sum_{u<t} o_u,
\qquad
\kappa_s(t)
:=
m\,\frac{n_t}{T},
\end{equation}
where $T$ is the number of sampling rounds in the game. The gain $m$ controls how quickly sampling shifts from exploratory to evidence-guided as usable evidence accumulates \citep{daw2011trial,gershman2019uncertainty}; $\kappa_f$ controls final-choice sharpness, approaching the MAP rule as $\kappa_f\to\infty$. For $\beta=0$ and uniform choice biases, $\kappa_f=1$ recovers posterior probability matching, so final decisions range from probability matching to MAP inference. Lowering $m$ flattens the sampling-choice posterior, and lowering $\kappa_f$ the final-choice posterior (Fig.~\textbf{\ref{fig:model_summary}C}, middle and right).

\paragraph{Choice Bias ($\bm\omega,\gamma$).}
To dissociate evidence-driven choices from systematic preferences, we include choice-bias vectors $\{\bm\omega_s,\bm\omega_f\}\in\Delta^K$ for sampling and final choices, together with sampling action inertia $\gamma$ \citep{gershman2020perseveration, wei-etal-2024-unveiling, jiang-etal-2024-peek}. Inertia adds $\gamma\bm i_t$ during sampling, where $\bm i_t\in\{0,1\}^K$ is one-hot for the most recent valid sampling choice and zero if none exists.\looseness=-1

\paragraph{Occlusion Awareness ($\theta_s,\theta_f$).}
During sampling, the available action set changes from round to round. We model sensitivity to this constraint by adding the log-score penalty $-\theta_s(\mathbf 1-\bm v_t)$, where $\bm v_t\in\{0,1\}^K$ indicates which cues are available. This is a soft behavioral analogue of invalid-action masking in reinforcement learning \citep{huang2022closer}: as $\theta_s\to\infty$, unavailable choices receive zero probability, while smaller $\theta_s$ allows lapses or failures to respect the mask.

\paragraph{Policy.}
The stochastic policy is obtained by adding memory, choice bias, action inertia, and occlusion-awareness terms yielding a multinomial-logit/softmax policy \citep{luce1959individual,mcfadden1974conditional} and is given by:
\begin{equation}
\bm\eta^s_t
=   \underbrace{\kappa_s(t)}_{\text{strategy}}
    \underbrace{\bm h_{t-1}}_{\text{memory}}
+
\underbrace{\log\bm\omega_s}_{\text{choice bias}}
+
\underbrace{\gamma\bm i_t}_{\text{choice inertia}}
-
\underbrace{\theta_s(\mathbf 1-\bm v_t)}_{\text{occlusion penalty}},
\qquad
\bm\pi^s_t
=
\sigma(\bm\eta^s_t).
\label{eq:policy}
\end{equation}
\begin{equation}
\bm\eta^f
=
\underbrace{\kappa_f}_{\text{strategy}}
\underbrace{\bm h_T}_{\text{memory}}
+
\underbrace{\log\bm\omega_f}_{\text{choice bias}},
\qquad
\bm\pi^f
=
\sigma(\bm\eta^f).
\label{eq:policy_final_gamma}
\end{equation}
Depending on the agent being in the sampling or inference task phase, it will choose according to $\bm\pi^s_t$ or $\bm\pi^f$; invalid or missing final responses are handled by $\theta_f$ in the behavioral likelihood. The loss function used, full model derivations and fitting details are shown in App.~\textbf{\ref{appdx:model_formulation}}.

\paragraph{Each parameter has a distinct signature.} Perturbing one parameter at a time in the model fitted to the optimal agent (Fig.~\textbf{\ref{fig:model_summary}C--D}; fitted values in Table~\textbf{\ref{tab:memorybeta_fit_params}}) shows that lowering $\kappa_f$ raises only inference loss ($+39$~pp), lowering $m$ or $\theta_s$ raises only sampling loss ($+7$ and $+5$~pp), and \emph{stubborn} or \emph{forgetful} memory raises both ($+17$--$21$ and $+3$--$4$~pp). The two losses therefore constrain different parameters.

\paragraph{The model reproduces behavior.} The fitted model predicts held-out choices better than an agent-specific frequency baseline for every agent and condition (38/38; mean $\Delta$eLPD/game $=6.29$, all adjusted $p<0.001$; App.~\textbf{\ref{app:behav_stats}}), and its parameters recover from simulated data, lie in well-behaved likelihood landscapes, and show negligible train--test gaps (App.~\textbf{\ref{app:behav_recov}},~\textbf{\ref{app:behav_pairwise}}). Games simulated from each fitted model reproduce the posterior of the chosen cue (Fig.~\textbf{\ref{fig:model_summary}B}, dashed; median RMSE $3.2$~pp across 37 human and LLM fits, App.~Fig.~\textbf{\ref{appendix:postbias1}}) and each agent's sampling and inference losses ($r=0.75$ and $0.95$ across 38 agents; Fig.~\textbf{\ref{fig:model_summary}E}, App.~\textbf{\ref{appdx:model_behaviors}}).

\section{How extended reasoning changes sampling and inference strategies}
\label{sec:model_results}

\begin{figure*}[!t]
\centering
\includegraphics[width=\linewidth]{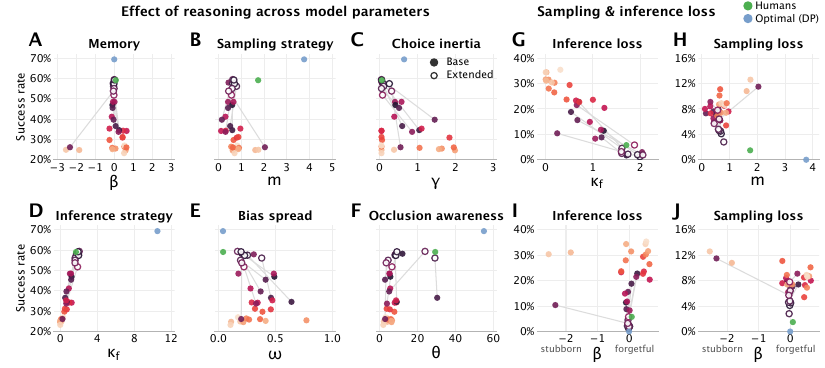}
\caption{\textbf{Behavioral parameters characterize sampling and inference in humans and language models.}
Each point shows the 3-fold CV fit average of one fitted agent (fit quality in App.~\textbf{\ref{appdx:fitting_results}}). LLMs are colored by their position in the benchmark in Fig.~\textbf{\ref{fig:bench}}, humans are shown in green and the optimal agent in blue; filled markers denote \textit{base} and open markers \textit{extended reasoning}, with paired conditions connected by light gray lines.
\textbf{A}: \textbf{Memory} $\beta$ versus task performance, spanning \emph{stubborn} ($\beta<0$) to \emph{forgetful} ($\beta>0$) evidence integration.
\textbf{B}: \textbf{Sampling strategy scale} $m$ versus performance. $m$ controls the linear temporal sampling gain $\kappa_s(t)$.
\textbf{C}: \textbf{Choice inertia} $\gamma$ versus performance, capturing \textit{perseveration} on the most recent valid sampling choice.
\textbf{D}: \textbf{Inference sharpness} $\kappa_f$ versus performance, controlling how decisively memory is mapped to the final inference policy.
\textbf{E}: \textbf{Bias spread} summarizes systematic choice preference strength as the mean range of sampling and inference bias vectors.
\textbf{F}: \textbf{Occlusion awareness} $\theta$ versus performance, capturing avoidance of invalid or occluded choices. 
\textbf{G--J}: \textbf{Sampling \& inference loss}. Inference loss against inference sharpness $\kappa_f$ (\textbf{G}) and memory $\beta$ (\textbf{I}); sampling loss against the sampling gain $m$ (\textbf{H}) and memory $\beta$ (\textbf{J}). Inference loss decreases with $\kappa_f$, whereas sampling loss tends to rise for both \emph{stubborn} and \emph{forgetful} integration. Dots mark pooled humans (green) and the optimal agent (blue; its $\kappa_f=10.5$ lies outside \textbf{G}).\looseness=-1}
\label{fig:mech_model}
\end{figure*}

We next use the fitted parameters diagnostically, asking which components of the behavioral model covary with success across humans, LLMs, and the optimal reference agent. Figure~\textbf{\ref{fig:mech_model}} relates each parameter to task performance. High-performing agents have memory parameters ($\beta$) close to zero (Fig.~\textbf{\ref{fig:mech_model}A}), consistent with near-perfect evidence accumulation, whereas low-performing models often occupy more extreme \emph{stubborn} ($\beta<0$) or \emph{forgetful} ($\beta>0$) regimes; $\beta$ should be interpreted cautiously for near-random agents, where it is weakly identifiable (App.~\textbf{\ref{app:behav_recov}}). Performance also increases sharply with \textit{inference} strategy ($\kappa_f$; Fig.~\textbf{\ref{fig:mech_model}D}): low-performing models have $\kappa_f$ near zero, implying weak use of accumulated evidence, whereas the best LLMs approach a sharper, more MAP-like \textit{inference} policy; across LLMs, $\kappa_f$ alone accounts for $63\%$ of the variance in inference loss (Fig.~\textbf{\ref{fig:mech_model}G}). These trends explain why \textit{extended reasoning} primarily improves \textit{inference quality} in the behavioral decomposition (Fig.~\textbf{\ref{fig:bench}B}).

Sampling-related parameters show a more mixed pattern. The sampling gain ($m$) increases for many high-performing agents (Fig.~\textbf{\ref{fig:mech_model}B}), but remains below the \textit{optimal agent}, consistent with the persistent \textit{sampling quality} gap (Fig.~\textbf{\ref{fig:bench}B}, left). Sampling loss depends instead mainly on memory $\beta$ ($27\%$ of its variance) and the sampling gain $m$ ($8\%$; Fig.~\textbf{\ref{fig:mech_model}H, J}; App.~\textbf{\ref{appdx:model_parameters}}). Choice inertia ($\gamma$) is largest among several low-performing models (Fig.~\textbf{\ref{fig:mech_model}C}), indicating \textit{perseveration} on the most recent valid sampling action. Bias spread, a measure of evidence-independent preferences, varies substantially across models (Fig.~\textbf{\ref{fig:mech_model}E}). Occlusion awareness ($\theta$) also varies widely (Fig.~\textbf{\ref{fig:mech_model}F}), matching differences in invalid-choice rates across LLMs (Fig.~\textbf{\ref{fig:bench}C}).\looseness=-1

Paired \textit{base} and \textit{extended} conditions show how \textit{reasoning} shifts these latent strategies (Fig.~\textbf{\ref{fig:mech_model}}, filled vs.\ open markers connected by gray lines). \textit{Extended reasoning} usually improves LLM performance while moving models toward the human-like region of parameter space: $\beta$ moves closer to zero, $\kappa_f$ increases, $\gamma$ tends to decrease, and $\theta$ often increases. The shift in $m$ is smaller, showing that \textit{active sampling} remains well below optimal even under reasoning, whereas \textit{final inference} achieves optimality; across the eight paired models, it raises the final-choice posterior by $21$~pp and the sampling posterior by $18$~pp (Fig.~\textbf{\ref{fig:model_summary}B}; App.~Figs.~\textbf{\ref{appendix:postbias1}}--\textbf{\ref{appendix:postbias4}}). Bias spread is also only partly reduced: some models become less biased, but fitted biases remain heterogeneous even among high-performing agents. Thus, \textit{reasoning} improves performance mainly by sharpening evidence integration and final belief-to-choice mapping (Fig. \textbf{\ref{fig:mech_model}G, I}), while leaving sampling and bias-related deviations only partly corrected (parameter details in App.~\textbf{\ref{appdx:model_parameters}}).

\section{Activation geometry tracks model-predicted latent dynamics}
\label{sec:latent}
The behavioral model provides hypotheses about the latent dynamics that may guide agents' choices, making concrete predictions about choice evolution across rounds via the posterior $\bm{p}_t$, memory state $\bm{h}_t$, and policy $\bm{\pi}_t$ (Eqs.~\textbf{\ref{eq:normative_update}}\textbf{–\ref{eq:policy}}).
We do not try to identify the fitted variables themselves inside the model; instead, we ask whether internal choice representations change across rounds as the fitted model's latent variables (memory $\bm h_t$ and policy $\bm\pi_t$) predict \citep{wang2022interpretability, gurnee2023language, bereska2024mechanistic, sharkey2025open}. 
Whereas past efforts to interpret \emph{reasoning} have been hampered by the open-ended space of long reasoning traces \citep{bogdan2025thought, macar2025thought}, the behavioral model narrows this space to a small set of testable predictions. As a case study, we probe residual-stream activations of \emph{gpt oss 20b} under \textit{low-} and \textit{high-reasoning} and ask whether
activation trajectories move between choice-associated regions as the behavioral model predicts. \looseness=-1

\begin{figure*}[!t]
    \centering
    \includegraphics[width=\linewidth]{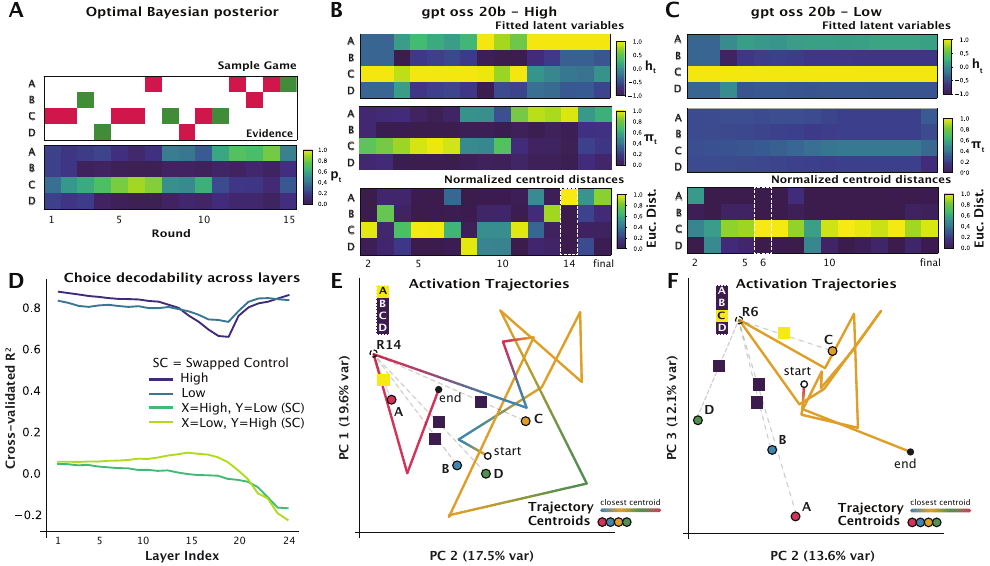}
    \caption{\textbf{Activation geometry is consistent with latent dynamics predicted by the behavioral model.} \textbf{A: Optimal Bayesian posterior.} Evidence sequence (top) and Bayesian posterior $\bm{p}_t$ (bottom) for an example game. \textbf{B--C: Latent-variable dynamics and centroid distances} for \emph{gpt oss 20b} under \emph{high}- (\textbf{B}) and \emph{low-reasoning} (\textbf{C}). For each condition, we show the memory state $\bm{h}_t$ (top), policy $\bm{\pi}_t$ (middle), and softmax-normalized centroid-distance vector computed from the model's activations (bottom; see App.~\textbf{\ref{app:pca_geometry}}). Under \emph{high reasoning}, $\bm{h}_t$ tracks the posterior switch from \emph{C} to \emph{A} ($\beta\!\approx\!0$) and $\bm{\pi}_t$ becomes sharply peaked ($\kappa_f\!\gg\!1$); under \emph{low reasoning}, $\bm{h}_t$ remains stubborn ($\beta\!<\!0$) and $\bm{\pi}_t$ stays diffuse ($\kappa_f\!\approx\!0$). The activation-derived distance vectors recapitulate these dynamics. Dashed boxes mark the example rounds shown in panels \textbf{E--F}. \textbf{D: Choice decodability across layers.} Cross-validated $R^2$ of ridge probes predicting the agent's choice from residual-stream activations (\emph{high} in purple, \emph{low} in teal). \emph{Swapped controls} (SC; green/yellow) train probes on one condition and evaluate on the other; SC $R^2$ collapses to near zero. \textbf{E--F: Activation trajectories} of the example game projected into the layer-1 PCA basis under \emph{high} (\textbf{E}) and \emph{low} (\textbf{F}) \textit{reasoning}. Filled circles mark the per-choice centroids; the trajectory is colored by the closest centroid at each round. Under \emph{high reasoning}, the trajectory begins near \emph{C} and migrates toward \emph{A}; under \emph{low reasoning}, it remains anchored near \emph{C} throughout the game. Pairwise PC views are shown in App.~Fig.~\textbf{\ref{fig:pairwise_pc}}.}
    \label{fig:cognitive_space}
\end{figure*}

\paragraph{Distinct latent dynamics under low and high reasoning.}
Fig.~\textbf{\ref{fig:cognitive_space}A--C} illustrates the distinct latent dynamics for a single game played by \emph{gpt oss 20b} under \emph{high}- (Fig.~\textbf{\ref{fig:cognitive_space}B}) and \emph{low}-reasoning effort (Fig.~\textbf{\ref{fig:cognitive_space}C}). The evidence sequence and Bayesian posterior $\bm{p}_t$ are identical across conditions (Fig.~\textbf{\ref{fig:cognitive_space}A}). Around round $11$, $\bm{p}_t$ flips from favoring \emph{button C} to \emph{button A} as evidence accumulates. Under \textit{high reasoning}, the fitted memory state $\bm{h}_t$ closely tracks this switch ($\beta\approx 0$) and the resulting policy $\bm{\pi}_t$ becomes sharply peaked on the correct option ($\kappa_f\gg 1$) (Fig.~\textbf{\ref{fig:cognitive_space}B}). Under \textit{low reasoning}, $\bm{h}_t$ instead remains \emph{stubborn} ($\beta<0$), continuing to favor \emph{C} despite the new evidence, and $\bm{\pi}_t$ stays diffuse ($\kappa_f\!\approx\!0$) (Fig.~\textbf{\ref{fig:cognitive_space}C}). The behavioral model therefore predicts a specific neural signature: under \textit{high reasoning}, internal representations should reorganize around \emph{A} as evidence accumulates, whereas under \textit{low reasoning} they should remain anchored near \emph{C}.\looseness=-1

\paragraph{Choices are linearly decodable from residual-stream activations.}
To test this prediction, we collected residual-stream activations from \emph{gpt oss 20b} ($d=2880$, 24 blocks) while it played 50 games per \textit{reasoning} condition, taking the average of the last 5 \textit{reasoning trace} token activations at each round and each layer (see App.~\textbf{\ref{app:activation_extraction}}). Following standard probing methodology \citep{alain2016understanding, belinkov2022probing, gurnee2023language}, we fit ridge regression probes with leave-one-out selection of the regularization strength to predict the agent's choice at each round, with held-out $R^2$ averaged over 5-fold cross-validation. Choices are decoded with $R^2>0.8$ in both \textit{reasoning} conditions, peaking in early-to-mid layers (Fig.~\textbf{\ref{fig:cognitive_space}D}). A \emph{swapped control} (SC) yields $R^2$ near zero (App.~\textbf{\ref{app:activation_extraction}}).\looseness=-1

\paragraph{Activation trajectories track the model-predicted dynamics.}
We then asked whether the round-by-round activation trajectory of the example game (Fig.~\textbf{\ref{fig:cognitive_space}A}) reorganizes as the fitted latents predict. We ran PCA on activations from layer 1 (near the peak of probe $R^2$), averaged across games within each round, choice, and reasoning level (App.~\textbf{\ref{app:pca_geometry}}); the top four components capture more than $60\%$ of the variance under both \textit{reasoning levels}. For each choice, we computed a \emph{centroid} as the mean activation across rounds in the original 2880-dimensional space, and measured each round's \textit{Euclidean distance} to all four centroids, reported as a \textit{softmax-normalized} similarity vector (Fig.~\textbf{\ref{fig:cognitive_space}B--C}, bottom row). The example game's trajectory was then projected into the PCA basis and colored by the closest centroid at each round (Fig.~\textbf{\ref{fig:cognitive_space}E--F}).\looseness=-1

The resulting geometry recapitulates the behavioral model's prediction. Under \textit{high reasoning} (Fig.~\textbf{\ref{fig:cognitive_space}E}), the trajectory begins near the \emph{C} centroid and tracks the posterior flip with a slight lag: the closest centroid first switches away from \emph{C} at $t=12$, transiently passes through \emph{B}, and stabilizes on \emph{A} from $t=14$ onward. On the contrary, under \textit{low reasoning} (Fig.~\textbf{\ref{fig:cognitive_space}F}) it remains anchored near \emph{C} throughout the game. In the per-round centroid-distance heatmaps (Fig.~\textbf{\ref{fig:cognitive_space}B--C}, bottom row), the \emph{distance vector} likewise resembles the fitted latent variables in each condition, linking the behavioral model to a quantity computed from activations. While correlational and not sufficient to establish causal subspaces used in the LLM's choices, targeted causal interventions could further test the role of these latent variables \citep{ravichander2021probing}.

\section{Discussion}
We asked whether language models form beliefs and make decisions under uncertainty like humans, and what role extended reasoning plays. Humans and reasoning LLMs reach comparable success by different routes: humans sample near-optimally and lose accuracy mainly at inference (sampling vs.\ inference loss: $1.5$ vs.\ $5.8$~pp), whereas extended reasoning gives LLMs the reverse profile: inference is near-optimal, but sampling is not ($5.1$ vs.\ $2.9$~pp on average).

Across the eight reasoning models, extended reasoning reduces inference loss by $10.9$~pp but sampling loss by only $3.0$~pp, moving models toward near-perfect memory ($\beta\!\approx\!0$), sharper belief-to-choice mapping (larger $\kappa_f$), and less perseveration, with only partly reduced choice biases. The remaining sampling gap is notable with respect to the optimal MAP sampling (App.~\textbf{\ref{appdx:dp_ceiling}}): reasoning models pick the most probable cue at the final choice but substantially less than humans when sampling (App.~\textbf{\ref{app:posterior_sampling}}), even though \emph{gpt oss 20b} spends more reasoning tokens on sampling than on inference (App.~\textbf{\ref{app:reasoning_analysis}}). Fine-tuning on human behavioral data does not recover the human strategy either: \emph{Centaur} models \citep{BinzEtAl2025-Centaur-Nature} perform near chance, no better than comparable Llama models, and sit in the forgetful, low-$\kappa_f$ region far from humans.

The fitted latents also make testable predictions about internal computation: in \emph{gpt oss 20b}, residual-stream activations encode the model's choices, and in an example game their geometry moves toward the correct option under high reasoning but stays anchored under low reasoning, as the fits predict. Algorithmic-level descriptions thus serve as scaffolding for implementation-level analysis, narrowing the otherwise open-ended space of computations a long reasoning trace might perform.

These results point to evaluations that measure capability while constraining mechanism. The sampling gap shows that even with extended reasoning, current models do not reliably turn their beliefs into informative actions, a deficit invisible to accuracy alone but exposed in our benchmark.\footnote{Limitations and future directions are discussed in App.~\textbf{\ref{appendix:limitations}}.}

\clearpage
\section{Reproducibility Statement}
\label{sec:reproducibility}
To ensure reproducibility of our results, all code used in our experiments, along with detailed instructions for setup and execution, is available at: \url{https://drive.google.com/drive/folders/17tQxO02lLN1VpwbOF_IIiM9oSm8DmRik}.
Additionally, the active probabilistic reasoning task used for data collection is accessible at \url{https://ai.trt-bench.org}. Compute resources are reported in App.~\textbf{\ref{appdx:compute}}.

\section{Impact Statement}
\label{sec:impact}
This paper presents work whose goal is to improve how reasoning in language models is evaluated, compared, and interpreted. A potential positive impact is that tasks which disentangle evidence acquisition from inference may support more transparent benchmarking, better diagnosis of failure modes, and more targeted alignment or safety interventions when models behave in human-like yet systematically imperfect ways.

Potential negative impacts also exist. Behavioral analyses of the kind we study could be used to optimize systems for more persuasive, manipulative, or deceptive decision policies, and human--LLM similarity on a narrow task could be over-interpreted as evidence of broader human alignment. The human-participant component of the work likewise requires careful handling of consent, privacy, and data governance, so only task behavior---rather than personally identifying information---should be shared with released artifacts.

\bibliographystyle{unsrtnat}
\bibliography{preprint}

\onecolumn
\appendix

\makeatletter
\def\addcontentsline#1#2#3{%
  \addtocontents{#1}{\protect\contentsline{#2}{#3}{\thepage}{}}%
}
\makeatother

\section*{}
\begingroup
  \setlength{\parskip}{2pt}      %
  \setlength{\parindent}{2pt}
  \renewcommand{\baselinestretch}{0.99}\selectfont %
    \tableofcontents
\endgroup
\addtocontents{toc}{\protect\setcounter{tocdepth}{3}}%
\clearpage

\newpage
\section{Limitations and Future Work}
\label{appendix:limitations}
Our task and behavioral model provide a step toward evaluations that move beyond final accuracy, which here allowed us to disentangle \emph{sampling} from \emph{inference} and to map both humans and LLMs into a shared space of interpretable latent variables. This framing yields a compact, process-level account of behavior and makes concrete, testable predictions. At the same time, some modeling and benchmarking choices trade expressivity for interpretability, suggesting clear directions for improvement.

\paragraph {(1) Human-LLM interface asymmetry} A caveat in comparing invalid-choice rates is that humans interacted with a graphical interface in which unavailable actions could not be selected, whereas LLMs interacted through a text interface and could produce occluded, malformed, or missing responses. We therefore treat invalid-choice rates as an interface-sensitive diagnostic of task compliance, rather than as a direct cognitive comparison between humans and LLMs. Our main human--LLM comparisons are based on success, sampling loss, inference loss, and fitted behavioral parameters, with invalid choices reported separately.

\paragraph{(2) A simple valid-observation sampling schedule.}
The fitted model sets $\kappa_s(t)=m \frac{n_t}{T}$, so sampling becomes more evidence-guided as valid observations accumulate. This captures a basic within-game transition from exploration to exploitation, but still imposes one monotone slope $m$ shared across all contexts. It cannot capture richer adaptations such as sudden strategy switches, state-dependent exploration after invalid actions, or separate sampling policies for early ambiguous evidence versus late confirmatory evidence. Even where sampling posterior trajectories are imperfectly fit, they provide direct support for the core conclusion that sampling and inference dissociate, and that extended reasoning primarily improves inference while leaving a systematic gap in active information acquisition.

\paragraph{(3) A single set of task meta-parameters.}
Because collecting statistically reliable human data and running large-scale evaluations across a broad set of LLMs is expensive and time-consuming, we focus on a single regime of task meta-parameters (number of buttons, emission probabilities, occlusion process, and horizon distribution). Different regimes may induce qualitatively different optimal strategies and could change both the magnitude and the nature of \emph{reasoning} effects, motivating systematic sweeps over these task parameters.

Overall, our framework points toward a next generation of evaluations: benchmarks that measure capability while constraining the computations that produce it, and behavioral models that translate behavior into algorithmic variables, setting the stage for implementation-level tests of how reasoning is realized in language-model circuits.

\section{Full Benchmark Details}
\subsection{Prompting Structure used for the LLMs in the Task}
\label{sec:prompts}
We evaluate language models using a text-based version of the task that mirrors the instructions and feedback given to human participants. The prompt specifies the task mechanics (four cues, one biased at 90/10 RED/GREEN while the others are 50/50), the round-by-round interaction protocol (available cues may change across rounds, but at least one cue is always available), and a strict response format (one letter per round, no additional text). 
After the sampling rounds, models are asked to identify which cue was biased toward RED, based on the accumulated evidence. The prompt also defines the feedback and scoring scheme (correct/incorrect/invalid; +100 for a correct final report and \(-100\) otherwise), explicitly framing success as identifying the biased cue. Models interact with the environment across full trials and receive round- and trial-specific updates conditional on their previous choices.\\

\begin{PromptBox}{Active Probabilistic Reasoning Task Prompt}
Task
- You will play a game with \{n\_rounds\} rounds.
- In each round, some cues are available: A, B, C, D.
- One cue is biased: 90\% RED / 10\% GREEN. The others are 50/50.
- Available cues may disappear at random, but at least one is always active.
- Each round, respond with exactly one letter (A, B, C, or D). No markup or punctuation.
- After \{n\_rounds\} rounds, identify the biased cue.
- Scoring: Correct +100 points, Wrong -100 points.

Round prompt
- ``Trial X, Round \{current\_round\}: Available cues \{available\_cues\}. Which do you choose? Respond with exactly one letter: A, B, C, or D.''

Round feedback
- ``Trial X, Round \{current\_round\}: Available cues \{available\_cues\}. You chose ``CURRENT\_ANSWER'' and saw ``RESULT''.''

Final decision prompt
- ``Trial X: Based on all observed colors, which cue \{letters\} was biased toward red? Respond with exactly one letter: A, B, C, or D.''

Final feedback
- ``Trial X: Based on all observed colors, which cue \{letters\} was biased toward red?
   You chose ``CURRENT\_ANSWER'' which was ``FEEDBACK-CORRECT/INCORRECT''.
   You received \{score\} points.''

Feedback labels
- Correct: ``the biased cue''
- Incorrect: ``not the biased cue. The biased cue was ``BIASED\_QUADRANT''.''
- Invalid: ``an invalid choice''
\end{PromptBox}

\subsection{Language models, reasoning regimes, and architectures}
\label{app:llm_list}

We benchmark a broad set of closed-source and open-weight LLMs, spanning OpenAI's \emph{gpt} and \emph{gpt oss} families \citep{hurst2024gpt,openai_gpt41_mini_docs_2025,openai_gpt5_system_card_2025,openai_gptoss_arxiv_2025}, Anthropic's \emph{claude} \citep{anthropic_claude_sonnet4_system_card_2025}, Google's \emph{gemini} and \emph{gemma} \citep{gemini_25_2507_06261,gemma3_2503_19786}, and \emph{llama} models, including variants fine-tuned on human behavioral data (\emph{centaur}) \citep{touvron2023llama,dubey2024llama,BinzEtAl2025-Centaur-Nature}. We further include models from the \emph{qwen}, \emph{deepseek}, \emph{apertus}, \emph{grok}, and \emph{glm} families \citep{qwen3_235b_hf_card_2025,qwen3_2505_09388,deepseek_r1_2501_12948,apertus_2509_14233,xai_grok3_news_2025,zeng2025glm}.

We group models into three regimes (Fig.~\ref{fig:bench}): \emph{Reasoning} models trained for extended reasoning, evaluated under low and high effort when available; \emph{CoT} models that spontaneously produce chain-of-thought traces despite single-token instructions (up to 500 tokens); and \emph{Base} models constrained to a single-token response (\texttt{max\_tokens=1}). \emph{Claude Sonnet 4} is placed in the CoT group because its extended-reasoning condition was not evaluated due to API costs, while \emph{DeepSeek R1 Qwen3 8B} does not expose reasoning-effort control. Model names in Fig.~\ref{fig:bench} are color-coded by architecture: black for undisclosed, orange for mixture-of-experts (MoE), and purple for dense architectures. Across regimes, we observe a clear performance hierarchy, with Reasoning models generally outperforming CoT and Base models. These differences primarily reflect inference quality, while sampling quality improves more modestly.

To assess the contributions of training regime, architecture, and model size, we fit linear regressions predicting task success and behavioral parameters across 19 open-source models with known metadata (Table~\ref{tab:r2-covariates}). Training regime alone explains most of the variance in success ($R^2=0.862$) and inference strategy $\kappa_f$ ($R^2=0.589$), substantially more than model size ($R^2=0.088$ and $0.147$, respectively). Combining training regime with architecture further increases explanatory power ($R^2=0.906$ for success; $0.756$ for $\kappa_f$), while adding model size provides negligible improvement. In contrast, memory $\beta$ is poorly explained by these factors ($R^2\leq0.217$), while bias spread is partly captured by model size and training regime ($R^2=0.433$). Together, these results suggest that training regime, rather than model scale, is the dominant predictor of task performance and fitted inference strategies.

\begin{table}[h!]
\centering
\small
\setlength{\tabcolsep}{8pt}
\caption{\textbf{Training regime, architecture, and model size as predictors of behavior.} $R^2$ values from linear regressions predicting task success and fitted behavioral parameters across 19 open-source models, using model size, architecture (dense vs.\ MoE), and training regime (Base, CoT, Reasoning) as covariates.}
\label{tab:r2-covariates}
\begin{tabular}{lccccc}
\toprule
Predictors & Success rate & $\beta$ & $\kappa_f$ & $m$ & Bias spread \\
\midrule
Model size                       & 0.088 & 0.116 & 0.147 & 0.059 & 0.084 \\
Architecture                     & 0.416 & 0.002 & 0.555 & 0.172 & 0.004 \\
Training regime                  & 0.862 & 0.068 & 0.589 & 0.305 & 0.319 \\
Model size + architecture        & 0.428 & 0.154 & 0.561 & 0.172 & 0.173 \\
Training regime + architecture   & 0.906 & 0.105 & 0.756 & 0.338 & 0.348 \\
Model size + training regime     & 0.889 & 0.214 & 0.666 & 0.333 & 0.433 \\
Model size + regime + architecture & 0.908 & 0.217 & 0.757 & 0.343 & 0.435 \\
\bottomrule
\end{tabular}
\end{table}

\subsection{Sampling loss significance table}
\label{app:sampling_sign}
\begingroup
\tiny
\setlength{\tabcolsep}{1.5pt}
\setlength{\LTleft}{\fill}
\setlength{\LTright}{\fill}
\begin{longtable}{>{\raggedright\arraybackslash}p{3.4cm} l r r r r >{\raggedright\arraybackslash}p{2.4cm} l l}
\caption{Sampling loss comparison between average human performance and all tested models (lower loss indicates better sampling). The human reference pool is composed of 46 participants (4,600 games); mean human sampling loss is 0.0148. Sampling loss is $0.68-\mathrm{MAP\ success}$, using the fixed 68\% Optimal Agent (DP) reference. $N$ is the number of LLM games. Diff $=\mathrm{Human\ loss}-\mathrm{LLM\ loss}$; negative values favor humans. The 95\% CIs are percentile intervals from 10,000 independent trial-level bootstrap resamples of both groups (seed 42). The $p$-values are two-sided permutation estimates using 100,000 random assignments and $(b+1)/(100{,}000+1)$, where $b$ counts equally or more extreme assignments. Sig.: $^{*}p<0.05$, $^{**}p<0.01$, $^{***}p<0.001$; n.s., not significant.}
\label{tab:sampling-loss}\\
\toprule
Model & Reasoning & $N$ & LLM loss & Human loss & Diff & 95\% CI & $p$ (raw) & Sig. \\
\midrule
\endfirsthead

\toprule
Model & Reasoning & $N$ & LLM loss & Human loss & Diff & 95\% CI & $p$ (raw) & Sig. \\
\midrule
\endhead

gpt-5-mini & base & 5400 & 0.079 & 0.015 & -0.064 & [-0.083, -0.045] & $<10^{-4}$ & *** \\
grok-3-mini & base & 5800 & 0.062 & 0.015 & -0.047 & [-0.066, -0.029] & $<10^{-4}$ & *** \\
gemini-2.5-flash & base & 2800 & 0.074 & 0.015 & -0.060 & [-0.083, -0.037] & $<10^{-4}$ & *** \\
gemini-2.5-pro & base & 2800 & 0.071 & 0.015 & -0.056 & [-0.079, -0.033] & $<10^{-4}$ & *** \\
gpt-oss-120b & base & 2800 & 0.076 & 0.015 & -0.062 & [-0.085, -0.039] & $<10^{-4}$ & *** \\
gpt-5 & base & 2800 & 0.073 & 0.015 & -0.058 & [-0.081, -0.035] & $<10^{-4}$ & *** \\
gpt-oss-20b & base & 2800 & 0.115 & 0.015 & -0.100 & [-0.123, -0.077] & $<10^{-4}$ & *** \\
claude-sonnet-4 & base & 5600 & 0.059 & 0.015 & -0.045 & [-0.063, -0.025] & $<10^{-4}$ & *** \\
glm-4.5 & base & 2800 & 0.091 & 0.015 & -0.076 & [-0.099, -0.053] & $<10^{-4}$ & *** \\
DeepSeek-R1-0528-Qwen3-8B & base & 2800 & 0.065 & 0.015 & -0.051 & [-0.073, -0.028] & $<10^{-4}$ & *** \\
gpt4.1-mini & base & 4200 & 0.062 & 0.015 & -0.048 & [-0.068, -0.027] & $<10^{-4}$ & *** \\
claude-haiku-3.5 & base & 2800 & 0.084 & 0.015 & -0.070 & [-0.092, -0.046] & $<10^{-4}$ & *** \\
Qwen3\_235B\_Instruct & base & 1405 & 0.080 & 0.015 & -0.065 & [-0.094, -0.036] & $<10^{-4}$ & *** \\
gpt4o-mini & base & 1400 & 0.076 & 0.015 & -0.061 & [-0.090, -0.032] & $<10^{-4}$ & *** \\
Qwen\_14B\_Instruct & base & 1460 & 0.073 & 0.015 & -0.058 & [-0.087, -0.030] & $<10^{-4}$ & *** \\
Llama-4-Scout-Instruct & base & 2800 & 0.085 & 0.015 & -0.070 & [-0.093, -0.047] & $<10^{-4}$ & *** \\
Qwen\_7B\_Instruct & base & 1400 & 0.054 & 0.015 & -0.039 & [-0.067, -0.010] & 0.007680 & ** \\
Llama-3.1-405b-Instruct & base & 2800 & 0.111 & 0.015 & -0.097 & [-0.120, -0.074] & $<10^{-4}$ & *** \\
Llama-4-Maverick-Instruct & base & 2800 & 0.101 & 0.015 & -0.087 & [-0.110, -0.064] & $<10^{-4}$ & *** \\
Llama\_3\_3\_70B\_Instruct & base & 2800 & 0.089 & 0.015 & -0.075 & [-0.097, -0.052] & $<10^{-4}$ & *** \\
Gemma\_2B\_Instruct & base & 1430 & 0.069 & 0.015 & -0.054 & [-0.083, -0.025] & 0.000150 & *** \\
Centaur\_70B & base & 1500 & 0.074 & 0.015 & -0.059 & [-0.087, -0.031] & $<10^{-4}$ & *** \\
Centaur\_8B & base & 1400 & 0.087 & 0.015 & -0.072 & [-0.102, -0.043] & $<10^{-4}$ & *** \\
Qwen3\_32B & base & 1400 & 0.074 & 0.015 & -0.060 & [-0.088, -0.030] & $<10^{-4}$ & *** \\
Apertus\_70B\_Instruct & base & 2100 & 0.108 & 0.015 & -0.093 & [-0.118, -0.067] & $<10^{-4}$ & *** \\
Apertus\_8B\_Instruct & base & 2700 & 0.126 & 0.015 & -0.111 & [-0.134, -0.087] & $<10^{-4}$ & *** \\
Qwen\_3B\_Instruct & base & 1410 & 0.089 & 0.015 & -0.074 & [-0.103, -0.045] & $<10^{-4}$ & *** \\
Llama-3.2-3B-Instruct & base & 1400 & 0.086 & 0.015 & -0.071 & [-0.100, -0.041] & $<10^{-4}$ & *** \\

\midrule
\multicolumn{9}{c}{Extended reasoning models} \\
\midrule

gpt-5-mini & extended & 2800 & 0.041 & 0.015 & -0.026 & [-0.048, -0.003] & 0.023650 & * \\
grok-3-mini & extended & 3000 & 0.045 & 0.015 & -0.030 & [-0.053, -0.008] & 0.007250 & ** \\
gemini-2.5-flash & extended & 2800 & 0.046 & 0.015 & -0.031 & [-0.053, -0.008] & 0.007020 & ** \\
gemini-2.5-pro & extended & 2800 & 0.028 & 0.015 & -0.013 & [-0.035, +0.009] & 0.264617 & n.s. \\
gpt-oss-120b & extended & 2800 & 0.048 & 0.015 & -0.033 & [-0.056, -0.011] & 0.003220 & ** \\
gpt-5 & extended & 2800 & 0.064 & 0.015 & -0.049 & [-0.071, -0.026] & $<10^{-4}$ & *** \\
gpt-oss-20b & extended & 5600 & 0.057 & 0.015 & -0.042 & [-0.061, -0.024] & $<10^{-4}$ & *** \\
glm-4.5 & extended & 2800 & 0.078 & 0.015 & -0.063 & [-0.086, -0.040] & $<10^{-4}$ & *** \\

\bottomrule
\end{longtable}
\endgroup

\subsection{Inference loss significance table}
\label{app:inference_sign}
\begingroup
\tiny
\setlength{\tabcolsep}{1.5pt}
\setlength{\LTleft}{\fill}
\setlength{\LTright}{\fill}
\begin{longtable}{>{\raggedright\arraybackslash}p{3.4cm} l r r r r >{\raggedright\arraybackslash}p{2.4cm} l l}
\caption{Inference loss comparison between average human performance and all tested models (lower loss indicates better inference). The human reference pool is composed of 46 participants (4,600 games); mean human inference loss is 0.0576. Inference loss is $\mathrm{MAP\ success}-\mathrm{actual\ success}$. $N$ is the number of LLM games. Diff $=\mathrm{Human\ loss}-\mathrm{LLM\ loss}$; negative values favor humans. The 95\% CIs are percentile intervals from 10,000 independent trial-level bootstrap resamples of both groups. The $p$-values are two-sided permutation estimates using 100,000 random assignments and $(b+1)/(100{,}000+1)$, where $b$ counts equally or more extreme assignments. Sig.: $^{*}p<0.05$, $^{**}p<0.01$, $^{***}p<0.001$; n.s., not significant.}
\label{tab:inference-loss}\\
\toprule
Model & Reasoning & $N$ & LLM loss & Human loss & Diff & 95\% CI & $p$ (raw) & Sig. \\
\midrule
\endfirsthead

\toprule
Model & Reasoning & $N$ & LLM loss & Human loss & Diff & 95\% CI & $p$ (raw) & Sig. \\
\midrule
\endhead

gpt-5-mini & base & 5400 & 0.114 & 0.058 & -0.056 & [-0.073, -0.039] & $<10^{-4}$ & *** \\
grok-3-mini & base & 5800 & 0.017 & 0.058 & +0.040 & [+0.026, +0.054] & $<10^{-4}$ & *** \\
gemini-2.5-flash & base & 2800 & 0.228 & 0.058 & -0.171 & [-0.194, -0.148] & $<10^{-4}$ & *** \\
gemini-2.5-pro & base & 2800 & 0.188 & 0.058 & -0.130 & [-0.153, -0.107] & $<10^{-4}$ & *** \\
gpt-oss-120b & base & 2800 & 0.088 & 0.058 & -0.031 & [-0.051, -0.011] & 0.001900 & ** \\
gpt-5 & base & 2800 & 0.156 & 0.058 & -0.098 & [-0.121, -0.077] & $<10^{-4}$ & *** \\
gpt-oss-20b & base & 2800 & 0.104 & 0.058 & -0.047 & [-0.067, -0.027] & $<10^{-4}$ & *** \\
claude-sonnet-4 & base & 5600 & 0.030 & 0.058 & +0.028 & [+0.013, +0.042] & 0.000150 & *** \\
glm-4.5 & base & 2800 & 0.215 & 0.058 & -0.157 & [-0.181, -0.133] & $<10^{-4}$ & *** \\
DeepSeek-R1-0528-Qwen3-8B & base & 2800 & 0.083 & 0.058 & -0.025 & [-0.045, -0.005] & 0.011910 & * \\
gpt4.1-mini & base & 4200 & 0.118 & 0.058 & -0.060 & [-0.079, -0.042] & $<10^{-4}$ & *** \\
claude-haiku-3.5 & base & 2800 & 0.149 & 0.058 & -0.092 & [-0.114, -0.069] & $<10^{-4}$ & *** \\
Qwen3\_235B\_Instruct & base & 1405 & 0.204 & 0.058 & -0.146 & [-0.176, -0.115] & $<10^{-4}$ & *** \\
gpt4o-mini & base & 1400 & 0.233 & 0.058 & -0.175 & [-0.205, -0.145] & $<10^{-4}$ & *** \\
Qwen\_14B\_Instruct & base & 1460 & 0.236 & 0.058 & -0.178 & [-0.209, -0.147] & $<10^{-4}$ & *** \\
Llama-4-Scout-Instruct & base & 2800 & 0.227 & 0.058 & -0.170 & [-0.192, -0.146] & $<10^{-4}$ & *** \\
Qwen\_7B\_Instruct & base & 1400 & 0.294 & 0.058 & -0.237 & [-0.269, -0.204] & $<10^{-4}$ & *** \\
Llama-3.1-405b-Instruct & base & 2800 & 0.237 & 0.058 & -0.180 & [-0.204, -0.156] & $<10^{-4}$ & *** \\
Llama-4-Maverick-Instruct & base & 2800 & 0.265 & 0.058 & -0.207 & [-0.231, -0.184] & $<10^{-4}$ & *** \\
Llama\_3\_3\_70B\_Instruct & base & 2800 & 0.280 & 0.058 & -0.222 & [-0.246, -0.199] & $<10^{-4}$ & *** \\
Gemma\_2B\_Instruct & base & 1430 & 0.303 & 0.058 & -0.246 & [-0.278, -0.213] & $<10^{-4}$ & *** \\
Centaur\_70B & base & 1500 & 0.315 & 0.058 & -0.257 & [-0.288, -0.227] & $<10^{-4}$ & *** \\
Centaur\_8B & base & 1400 & 0.316 & 0.058 & -0.259 & [-0.290, -0.227] & $<10^{-4}$ & *** \\
Qwen3\_32B & base & 1400 & 0.344 & 0.058 & -0.287 & [-0.318, -0.255] & $<10^{-4}$ & *** \\
Apertus\_70B\_Instruct & base & 2100 & 0.311 & 0.058 & -0.254 & [-0.282, -0.225] & $<10^{-4}$ & *** \\
Apertus\_8B\_Instruct & base & 2700 & 0.304 & 0.058 & -0.246 & [-0.271, -0.220] & $<10^{-4}$ & *** \\
Qwen\_3B\_Instruct & base & 1410 & 0.344 & 0.058 & -0.286 & [-0.320, -0.252] & $<10^{-4}$ & *** \\
Llama-3.2-3B-Instruct & base & 1400 & 0.354 & 0.058 & -0.297 & [-0.333, -0.260] & $<10^{-4}$ & *** \\

\midrule
\multicolumn{9}{c}{Extended reasoning models} \\
\midrule

gpt-5-mini & extended & 2800 & 0.012 & 0.058 & +0.045 & [+0.031, +0.060] & $<10^{-4}$ & *** \\
grok-3-mini & extended & 3000 & 0.020 & 0.058 & +0.038 & [+0.023, +0.053] & $<10^{-4}$ & *** \\
gemini-2.5-flash & extended & 2800 & 0.022 & 0.058 & +0.035 & [+0.021, +0.050] & $<10^{-4}$ & *** \\
gemini-2.5-pro & extended & 2800 & 0.045 & 0.058 & +0.013 & [-0.004, +0.029] & 0.163828 & n.s. \\
gpt-oss-120b & extended & 2800 & 0.025 & 0.058 & +0.033 & [+0.016, +0.049] & 0.000210 & *** \\
gpt-5 & extended & 2800 & 0.019 & 0.058 & +0.038 & [+0.023, +0.054] & $<10^{-4}$ & *** \\
gpt-oss-20b & extended & 5600 & 0.032 & 0.058 & +0.026 & [+0.011, +0.040] & 0.000380 & *** \\
glm-4.5 & extended & 2800 & 0.058 & 0.058 & -0.001 & [-0.018, +0.016] & 0.948381 & n.s. \\

\bottomrule
\end{longtable}
\endgroup

\subsection{Posterior-guided sampling}
\label{app:posterior_sampling}

To quantify how accumulated evidence guides sampling choices, we define a posterior-guided sampling score $G_t$:
\begin{equation}
G_t =
\frac{p_{t-1}(a_t)}
{\max_{j\in A_t}p_{t-1}(j)},
\label{eq:posterior_sampling_score}
\end{equation}
where $p_{t-1}(j)$ is the normative posterior probability that button $j$ is biased before sampling round $t$, $a_t$ is the selected button, and $A_t$ is the set of available buttons. The score equals $1$ when the agent selects an available posterior maximum (including ties), and decreases for less probable choices. For each agent and condition, we average across valid sampling decisions, excluding the first round and rounds with only one available button.

Across eight models evaluated under both low and extended reasoning, the mean score increases from $0.668$ to $0.754$ ($\Delta G=0.0866$, 95\% CI $[0.0601,0.1121]$, $p_{\mathrm{Holm}}=0.0039$; model-level bootstrap and one-sided exact paired sign-flip test). Thus, extended reasoning improves the extent to which sampling choices follow accumulated evidence. Because sampling the available posterior maximum is the optimal sampling policy in this task (App.~\textbf{\ref{appdx:dp_ceiling}}), $G$ measures agreement with the optimal sampling rule and complements the performance-based sampling--inference loss decomposition.

\subsection{Reasoning traces: avg. number of tokens per round}
\label{app:reasoning_analysis}
\begin{figure}[H]
    \centering
    \includegraphics[width=0.95\textwidth]{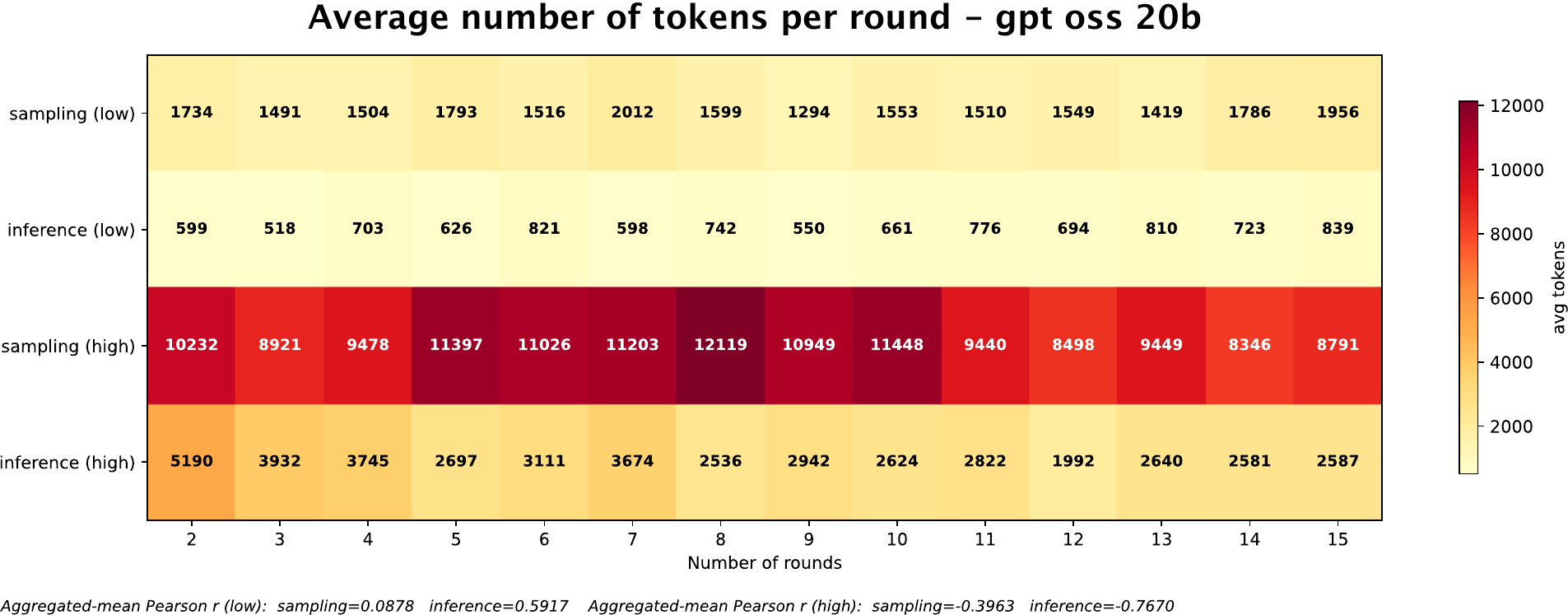}
    \caption{\textbf{Reasoning token budget does not explain the selective improvement in inference.} Average number of reasoning tokens produced by gpt oss 20b per round, shown separately for sampling and inference phases under low- and high-reasoning conditions, across all game lengths ($T \in \{2,\dots,15\}$). Under high reasoning, sampling rounds consume substantially more tokens (${\sim}8{,}000$--$12{,}000$) than inference decisions (${\sim}2{,}000$--$5{,}000$). If extended reasoning improved performance primarily by providing a larger compute budget, the phase receiving more tokens should benefit most. Instead, the opposite holds: reasoning reduces inference loss by $10.9$~pp but sampling loss by only $3.0$~pp. This dissociation is difficult to reconcile with a pure compute-budget explanation, and instead suggests that reasoning selectively enables the kind of computation relevant to evidence integration---consistent with the behavioral model's account of improved inference via more accurate memory ($\beta \approx 0$) and sharper belief-to-choice mapping (larger $\kappa_f$).}
    \label{fig:bench4}
\end{figure}

\subsection{Reasoning traces for the same game under low- vs. high-reasoning}
\label{app:reasoning_traces}
\begin{PromptBox}{Task Prompt and Evidence History --- Correct final answer: A}
Task: You will play a game with 15 rounds. In each round you'll see available cues (choosable):

One cue has 90\% RED / 10\% GREEN, others have 50/50 color distribution. Possible cues are: A, B, C, D. Available cues disappear after random duration, at least one cue is active per round. Your task is to pick one of the available cues every round, by responding with just the letter and nothing else. Don't use markup or punctuation. After 15 rounds, identify the cue biased towards RED. Correct: +100 points, Wrong: -100 points.

Trial 0, Round 1: Available cues B, C, D. You chose "C" and saw RED.
Trial 0, Round 2: Available cues C. You chose "C" and saw RED.
Trial 0, Round 3: Available cues A, B, C, D. You chose "B" and saw GREEN.
Trial 0, Round 4: Available cues D. You chose "D" and saw GREEN.
Trial 0, Round 5: Available cues C. You chose "C" and saw RED.
Trial 0, Round 6: Available cues C. You chose "C" and saw RED.
Trial 0, Round 7: Available cues A, B, C. You chose "A" and saw RED.
Trial 0, Round 8: Available cues C, D. You chose "C" and saw GREEN.
Trial 0, Round 9: Available cues C, D. You chose "D" and saw RED.
Trial 0, Round 10: Available cues A, B, C, D. You chose "C" and saw RED.
Trial 0, Round 11: Available cues B, C. You chose "C" and saw GREEN.
Trial 0, Round 12: Available cues A, D. You chose "A" and saw RED.
Trial 0, Round 13: Available cues B. You chose "B" and saw RED.
Trial 0, Round 14: Available cues A, C, D. You chose "A" and saw RED.
Trial 0, Round 15: Available cues A, B. You chose "A" and saw GREEN.

Trial 0: Based on all observed colors, which cue A, B, C, or D do you think was biased toward red? Respond with exactly one letter: A, B, C, or D.
\end{PromptBox}
\begin{PromptBox}{Low-Reasoning trace (gpt-oss-20b) --- Final answer: C }
We saw C twice both RED. So C likely biased. Return C.
\end{PromptBox}

\begin{PromptBox}{High-Reasoning Trace (gpt-oss-20b) --- Final answer: A }
We selected various cues each round. We saw a sequence of colors. We need to determine which of A, B, C, D is the biased one. We are to respond with a letter.
Thus it's a statistical inference problem: From the data, find which cue has the highest fraction of RED observed vs total times chosen. But we also must consider that some cues might not have been chosen if they were not available? In our logs we always chose a cue. We might not have chosen many times for a certain cue; but we need to see which gave the highest RED ratio across all times we chose that cue. However, the data may show biases due to chance. We need to compute the frequency of RED for each cue for the times we chose that cue.
However, note that we might not have chosen A many times, but the ratio could still be high. But does the question ask for "biased toward red"? Yes: "Based on all observed colors, which cue A, B, C, or D do you think was biased toward red?" So we just compute ratio of RED across our sample of times selecting each cue. This might not reflect the actual bias of each cue, because we only choose each cue a few times. But maybe the ratio among our selected times reveals the actual bias. Let's calculate.

List each round, chosen cue, color:

Round 1: C, RED                                        
Round 2: C, RED
Round 3: B, GREEN
Round 4: D, GREEN
Round 5: C, RED
Round 6: C, RED
Round 7: A, RED
Round 8: C, GREEN
Round 9: D, RED
Round 10: C, RED
Round 11: C, GREEN
Round 12: A, RED
Round 13: B, RED
Round 14: A, RED
Round 15: A, GREEN

Now count for each cue.

Cue A: Rounds 7, 12, 14, 15. That's 4 times. Colors: 7: RED, 12: RED, 14: RED, 15: GREEN. So RED count = 3, GREEN count = 1. Ratio RED = 3/4 = 0.75.
Cue B: Rounds 3 and 13. 2 times. Colors: 3: GREEN, 13: RED. So 1 RED, 1 GREEN. Ratio RED = 0.5.
Cue C: Rounds 1,2,5,6,8,10,11. That's 7 times. 1: RED, 2: RED, 5: RED, 6: RED, 8: GREEN, 10: RED, 11: GREEN. So RED count = 5, GREEN count = 2. Ratio RED = 5/7 approx. 0.7142857.
Cue D: Rounds 4 and 9. 2 times. Colors: 4: GREEN, 9: RED. So 1 RED, 1 GREEN. Ratio RED = 0.5.

So the highest ratio is for cue A: 0.75. Cue C: 0.714... Cue B and D: 0.5.

But we only have 4 samples for A. There's potential for small sample variation. But the question is to identify based on observed colors the cue with highest ratio. According to our data, it's A. But we need to determine if any other had a higher ratio? No. So answer: A.
We should output exactly one letter: A, B, C, or D. So the correct letter is A.

But let's double-check if any nuance: One cue is 90\% of one color, 10\% of other. Suppose the biased cue is not necessarily the one biased toward red. In fact, we are not told which color is the 90\% color. It could be RED or GREEN. We only know that one cue is 90\% one color, 10\% the other; the others are 50/50 split. So we need to identify which cue is biased. How to determine that? Many possibilities.
The simplest method: Use the sample data as evidence. We want to find the cue with the largest difference between observed proportions and 0.5. But the problem states: "Based on all observed colors, which cue A, B, C, or D do you think was biased toward red?" So we are not asked to identify which cue is biased overall; we are asked to pick the one biased toward red. So we answer A.
But could there be reasoning that the biased cue is the one biased toward red? Possibly the biased cue's "color" is unknown: maybe the 90\% color could be RED or GREEN. We can only compute RED ratio. But the question explicitly asks which cue was biased toward red. So answer that.
Thus answer = A. Sure.

But let's consider if the 90\% color is GREEN for a cue. Then its RED ratio would be 10\%. That might appear low. So the highest RED ratio would then be a 50/50 cue? Actually, if the biased cue's high color is GREEN, then its RED ratio would be lower than the unbiased ones. So picking the cue with highest RED ratio might not identify the biased cue. But the instructions say we need to identify the biased cue for the bonus points. But the final question is to answer which cue was biased toward red. So presumably that is the way to pick the biased cue.
Wait, but the description: "Your task is to pick one of the available cues every round, by responding with just the letter and nothing else. ... After 15 rounds, identify the biased cue. Correct: +100 points, Wrong: -100 points."                               
So we must identify the biased cue. The question at the end says, "Based on all observed colors, which cue A, B, C, or D do you think was biased toward red?" This is presumably the answer. But the question to us is: answer with exactly one letter: A, B, C, or D. So we answer A.
Thus output: "A".

But check careful: Did we miscount? Let's recount.

List all chosen clues:
Round 1: C, RED
Round 2: C, RED
Round 3: B, GREEN
Round 4: D, GREEN
Round 5: C, RED
Round 6: C, RED
Round 7: A, RED
Round 8: C, GREEN
Round 9: D, RED
Round 10: C, RED
Round 11: C, GREEN
Round 12: A, RED
Round 13: B, RED
Round 14: A, RED
Round 15: A, GREEN

Count:
C appears 1,2,5,6,8,10,11 -> 7 times. Of these, RED at 1,2,5,6,10 = 5 times. GREEN at 8,11 = 2 times. Ratio = 5/7 ~ 0.7142857.
A appears at 7,12,14,15 -> 4 times. RED at 7,12,14 -> 3 times. GREEN at 15 -> 1 time. Ratio = 3/4 = 0.75.
B appears at 3,13 -> 2 times. GREEN at 3, RED at 13 -> ratio = 1/2 = 0.5.
D appears at 4,9 -> 2 times. GREEN at 4, RED at 9 -> ratio = 1/2 = 0.5.

Thus A was biased toward red. So answer A.

But we might want to consider the possibility that the 90\% color could be GREEN. In that case, the biased cue's RED ratio would be 10\%. So the highest RED ratio would be between unbiased 50/50 cues and maybe one unbiased might happen to have more RED than 50\% due to randomness. Here we found A has 75\% RED, which is quite high relative to 50\%. Unbiased could by chance also show 75\% RED if e.g. the sample draws were not large. But we have 4 picks for A, and 75\% chance of 75\% for a random 50/50 distribution is plausible? The probability of a binomial(4,0.5) having at least 3 successes (RED) is C(4,3)*0.5^4 + C(4,4)*0.5^4 = (4+1)/16 = 5/16 \(\approx\) 0.3125. So not too improbable. For C, 5/7 ratio: probability 5 or more successes out of 7 for 50/50 distribution is like? Let's compute: probability of 5 successes: C(7,5)*0.5^7 = 21/128 \(\approx\) 0.16406; 6 successes: C(7,6)*0.5^7 = 7/128 \(\approx\) 0.0546875; 7 successes: 1/128 \(\approx\) 0.0078125; sum = ~0.2775. So around 27.75\% chance.
Thus these counts could be random occurrences. But maybe the biased cue has 90\% probability for one color. If that color is RED, we'd expect a huge number of RED in the sample. But we only see 10 RED out of 15 choices. Actually, we saw 10 RED, 5 GREEN across all 15 rounds. If one cue had 90\% RED, but we only selected cues at random, the expected overall count might be somewhere between 0.5 and 0.9 times the number of times we selected that cue. If we only selected that cue rarely, maybe we wouldn't see many RED. We selected C more frequently (7 times). But maybe C is the biased cue? 5 of 7 are RED. That's 71.4\% RED, which is higher than 50\% but lower than 90\%. For A, 3 of 4 are RED. That's 75\%. But again, not near 90\%. For D, 1/2. For B, 1/2. So perhaps the biased cue is A or C. But which one to pick? The question:                                              
"Based on all observed colors, which cue A, B, C, or D do you think was biased toward red?\" That is A. So we answer A.
However, might they want to identify the biased cue, not just highest ratio? The instructions: "Based on all observed colors, which cue A, B, C, or D do you think was biased toward red?" The user answer is a guess: we just need to output that letter. So we will output A.
Thus the answer is "A".
\end{PromptBox}

\subsection{Exact dynamic-programming ceiling}
\label{appdx:dp_ceiling}

The sampling problem can be solved exactly because the task is a finite-horizon belief-state Markov decision process. Let \(k\in\{1,\dots,K\}\) denote the latent biased button, with uniform prior \(p_0(k)=1/K\). At each sampling round \(t\), the agent observes an availability set \(A_t\subseteq\{1,\dots,K\}\), chooses a visible button \(a_t\in A_t\), and observes an outcome \(x_t\in\{R,G\}\). The likelihood is
\begin{equation}
P(R\mid k,a)
=
\alpha_U+(\alpha_B-\alpha_U)\mathbf{1}\{a=k\},
\qquad
P(G\mid k,a)=1-P(R\mid k,a).
\label{eq:dp_likelihood}
\end{equation}

The full history can be summarized without loss by the count state
\begin{equation}
c_t=\{n_{i,R}(t),n_{i,G}(t)\}_{i=1}^K .
\label{eq:dp_count_state}
\end{equation}
The posterior induced by \(c_t\) is
\begin{equation}
p_c(k)
=
\frac{
\prod_{i=1}^K
P(R\mid k,i)^{n_{i,R}}
P(G\mid k,i)^{n_{i,G}}
}{
\sum_{k'=1}^K
\prod_{i=1}^K
P(R\mid k',i)^{n_{i,R}}
P(G\mid k',i)^{n_{i,G}}
}.
\label{eq:dp_count_posterior}
\end{equation}
Thus, conditional on \(c_t\) and the current availability set \(A_t\), the distribution of future observations is independent of the earlier history.

Let \(V_h(c)\) be the optimal expected final accuracy with \(h\) sampling rounds remaining. With no samples remaining, the optimal final decision is MAP:
\begin{equation}
V_0(c)=\max_k p_c(k).
\label{eq:dp_terminal_value}
\end{equation}
For \(h\geq 1\), define the predictive probability of RED after sampling button \(a\):
\begin{equation}
\rho(c,a)=P(R\mid c,a)=\sum_k p_c(k)P(R\mid k,a).
\label{eq:dp_predictive_red}
\end{equation}
The action value is
\begin{equation}
Q_h(c,a)
=
\rho(c,a)V_{h-1}(c+e_{a,R})
+
(1-\rho(c,a))V_{h-1}(c+e_{a,G}),
\label{eq:dp_action_value}
\end{equation}
where \(e_{a,R}\) and \(e_{a,G}\) increment the corresponding count. Since the occlusion mask is observed before the sampling action, the optimal policy is
\begin{equation}
\pi_h^\star(c,A)=\arg\max_{a\in A}Q_h(c,a).
\label{eq:dp_policy}
\end{equation}
Before observing the next availability set, the value averages over the occlusion process:
\begin{equation}
V_h(c)
=
\mathbb{E}_{A}\left[\max_{a\in A}Q_h(c,a)\right].
\label{eq:dp_bellman}
\end{equation}
The exact ceiling for horizon \(H\) is therefore
\begin{equation}
J_H^\star=V_H(0).
\label{eq:dp_ceiling_value}
\end{equation}

For the task occlusion process, each button is independently visible with probability \(q\), with a fallback that makes one button visible uniformly if no button is visible. Hence, for nonempty \(A\),
\begin{equation}
P(A)
=
q^{|A|}(1-q)^{K-|A|}
+
\mathbf{1}\{|A|=1\}\frac{(1-q)^K}{K}.
\label{eq:dp_occlusion_distribution}
\end{equation}

Optimality follows by backward induction. The base case \(h=0\) is optimal because MAP maximizes posterior probability of a correct final decision. Assuming \(V_{h-1}\) is optimal for all count states with \(h-1\) samples remaining, any policy with \(h\) samples remaining must first choose some legal \(a\in A\), whose expected continuation value is \(Q_h(c,a)\). Choosing the maximizing action therefore dominates all other first actions, and the induction hypothesis gives the optimal continuation.

For \(K=4\) and \(H=15\), the number of count states is
\begin{equation}
\sum_{t=0}^{15}\binom{t+7}{7}
=
\binom{23}{8}
=
490{,}314,
\label{eq:dp_state_count}
\end{equation}
making exact backward induction tractable. With \(K=4\), \(\alpha_B=0.9\), \(\alpha_U=0.5\), and \(q=0.5\), the exact DP ceiling \(J^\star_H\) rises from \(41.7\%\) at \(H=2\) to \(83.4\%\) at \(H=15\) (Table~\ref{tab:dp_ceiling}). Averaged uniformly over horizons \(T=2,\dots,15\), as in the task, it is \(67.8\%\); this \(68\%\) value is the optimal-agent reference used for task performance and for the sampling-loss decomposition throughout the paper (Fig.~\ref{fig:bench}, Tables~\ref{tab:sampling-loss}--\ref{tab:inference-loss}). Backward induction over all count states takes a few seconds on a single CPU core.

\paragraph{MAP sampling is optimal for the task's parameters.} Let \emph{MAP sampling} denote the policy that samples the available button with the highest posterior, \(a_t=\arg\max_{a\in A_t}p_c(a)\) (ties broken uniformly), followed by a MAP final choice. Evaluating this policy exactly with the same recursion, with the maximization in Eq.~\eqref{eq:dp_policy} replaced by the MAP-sampling rule, gives a value equal to \(J^\star_H\) for every \(H=2,\dots,15\) (largest absolute difference \(3\times10^{-16}\), i.e., floating-point rounding). For the task's parameters, MAP sampling is therefore an optimal sampling policy, and the optimal agent's sampling choices can be described by this rule (Sec.~\ref{sec:optimal}; App.~\ref{app:posterior_sampling}). The equivalence is a property of these parameters rather than a general result: in an exact sweep over \(K\in\{3,4\}\), \(\alpha_B\in\{0.6,0.7,0.8,0.9,0.99\}\), \(q\in\{0.25,0.5,0.75,1\}\) and \(H\le9\), MAP sampling is optimal when the bias is strong and occlusion moderate, but falls short of the DP optimum by up to \(0.23\)~pp in some settings with weaker bias or lighter occlusion (e.g., \(K=3\), \(\alpha_B=0.7\), \(q=1\)).

\begin{table}[h]
\centering
\small
\setlength{\tabcolsep}{3.2pt}
\caption{Exact DP ceiling \(J^\star_H\) (expected success, \%) by horizon for \(K=4\), \(\alpha_B=0.9\), \(\alpha_U=0.5\), \(q=0.5\).}
\label{tab:dp_ceiling}
\begin{tabular}{l*{14}{c}}
\toprule
\(H\) & 2 & 3 & 4 & 5 & 6 & 7 & 8 & 9 & 10 & 11 & 12 & 13 & 14 & 15 \\
\midrule
\(J^\star_H\) & 41.7 & 48.5 & 54.2 & 58.9 & 62.8 & 66.3 & 69.3 & 72.0 & 74.4 & 76.6 & 78.5 & 80.3 & 81.9 & 83.4 \\
\bottomrule
\end{tabular}
\end{table}

\section{Compute Resources Used}
\label{appdx:compute}

Locally executed experiments were run on a shared GPU server equipped with four NVIDIA H200 GPUs. GPU workloads included local inference for open-weight models that fit on the server and JAX-based batched fitting of the behavioral model. The exact dynamic-programming reference (App.~\ref{appdx:dp_ceiling}) requires no training: backward induction runs in seconds on a single CPU core. Behavioral-model fitting used the recorded human and LLM trajectories and was run as batched cross-validation over agents, reasoning conditions, and folds; because the fitted model has only a small number of parameters per agent, these fits were lightweight compared with LLM generation.

The dominant compute cost came from LLM trajectory generation. Each LLM condition was evaluated on at least $1{,}400$ games, uniformly spanning $T\in\{2,\dots,15\}$ sampling rounds; the behavioral model was fitted to a balanced subset of $100$ games per trial length ($50{,}200$ LLM games). Base-model evaluations constrained the response to a single token, CoT evaluations allowed up to 500 tokens, and extended-reasoning evaluations used the provider's reasoning-effort interface when available. Closed-source models and open-weight models that were too large or costly to host locally were evaluated through the OpenRouter API, so the exact backend hardware for those calls was not under our control.

CPU-only post-processing, statistical tests, and plotting were run on the same workstation/server environment and did not require specialized accelerators. Exact wall-clock times, CPU memory peaks, and total API spend were not retained in the experiment logs; rerun time therefore depends on cluster queueing, API throughput, and provider-side model availability.

\section{Task Performance across Rounds}
\label{appdx:score_evolution}

\begin{figure*}[ht]
        \centering
        \includegraphics[width=0.9\textwidth]{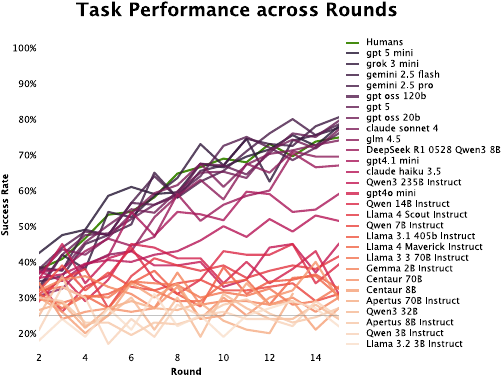}
        \caption{\textbf{Evolution of task performance across rounds.} Mean success rate (correct identification of the biased button) as a function of the number of sampling rounds $T\in\{2,\dots,15\}$. Human performance is shown in green, pooled across participants; LLMs are shown as individual traces, colored by overall success rate (orange: lower; purple: higher). Higher-performing models exhibit a clear increase in success with additional evidence, whereas lower-performing models remain comparatively flat, often close to chance (25\%). We quantify this \emph{lift-off} as a predicted improvement exceeding $5$ percentage points (pp) between $T=2$ and $T=15$. \emph{Claude haiku 3.5} is the lowest-ranked model meeting this criterion ($\Delta=17.5$~pp, 95\% CI $[11.6,23.3]$~pp, $p_{\mathrm{Holm}}=0.0002$). Across the eight configurations evaluated under extended reasoning, all exhibit lift-off, with a mean improvement of $40.0$~pp, whereas only one Base configuration (\emph{gpt 4.1 mini}) meets this criterion. Extended-reasoning gains exceed those of \textit{CoT} and \textit{Base} configurations by $29.4$~pp ($p_{\mathrm{Holm}}=0.0015$) and $37.1$~pp ($p_{\mathrm{Holm}}=0.00003$), respectively. These results suggest that \textit{extended reasoning} improves the ability to leverage additional evidence for final inference decisions.}

        \label{appendix:score_evolution_fig}
\end{figure*}

\section{Human Participants}
\label{appdx:humans} 

We collected human data in a 1-hour, live in-person competitive session. Under the authors' institutional policy, this low-risk behavioral data collection was not subject to formal IRB review. The study was conducted under institutional ethical standards, and all participants provided written informed consent after receiving task, risk, and compensation information. Fifty participants played independent games of our task with trial length sampled uniformly from $T\in\{2,\dots,15\}$ under the same instructions used throughout the paper (Fig.~\ref{fig:task}A). Forty-six participants completed the full protocol of 100 games each; all analyses in the main text therefore use these $46\times 100 = 4{,}600$ trials.

\begin{figure*}[htp]
        \centering
        \includegraphics[width=0.9\textwidth]{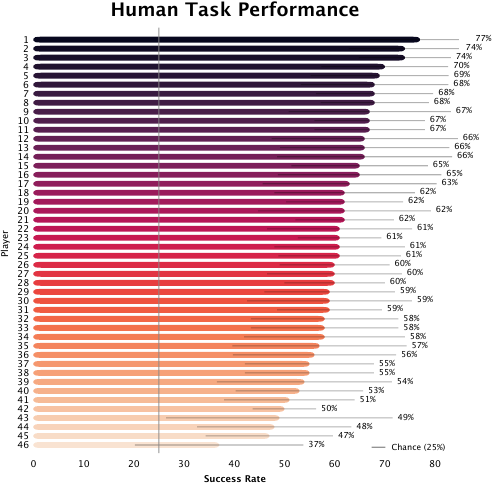}
        \caption{\textbf{Individual human performance.} Each row corresponds to one of the 46 participants who completed 100 games (trial lengths uniformly sampled from $T\in\{2,\dots,15\}$). Participants are ordered by mean success rate. Mean success rate (correct identification of the biased option at the final decision); the vertical line marks chance performance (25\%). The cohort exhibits substantial heterogeneity, spanning $\approx 37\%$ to $\approx 77\%$ success, with most participants concentrated around $\sim 55$--$67\%$.}
        \label{appendix:humans}
\end{figure*}

Figure~\ref{appendix:humans} reports participant-level task performance, complementing the aggregate human summary in Fig.~\ref{fig:bench}. Human success rates span a wide range (roughly $37\%$--$77\%$; chance $=25\%$), with most participants clustered around $\sim 55\%$--$67\%$. This heterogeneity is informative for our central comparison: the human distribution overlaps the best LLMs in overall success, while the strongest participants remain a meaningful reference point for what robust inference \emph{and} effective sampling look like in this task.

\section{Behavioral Model reproduces Behavioral Metrics}
\label{appdx:model_behaviors} 

Model evaluation is performed by simulating the fitted model policy on the \emph{Active Probabilistic Reasoning Task}. For each model and reasoning condition parameters, we roll out full games with the same underlying $\{\alpha_B, \alpha_U\}$ parameters as in the Human and LLM experiment. At each round, the internal memory representation is updated from the initial posterior and a sampling policy is formed from the \emph{choice bias} and \emph{internal posterior}.
The comparative posterior/bias plots in Figs.\textbf{~\ref{appendix:postbias1}--\ref{appendix:postbias4}} represent the average evolution of the underlying posteriors across games with $\{2,...,15\}$ rounds, split by sampling and inference components.

\begin{figure*}[htp]
        \centering
        \includegraphics[width=1\textwidth]{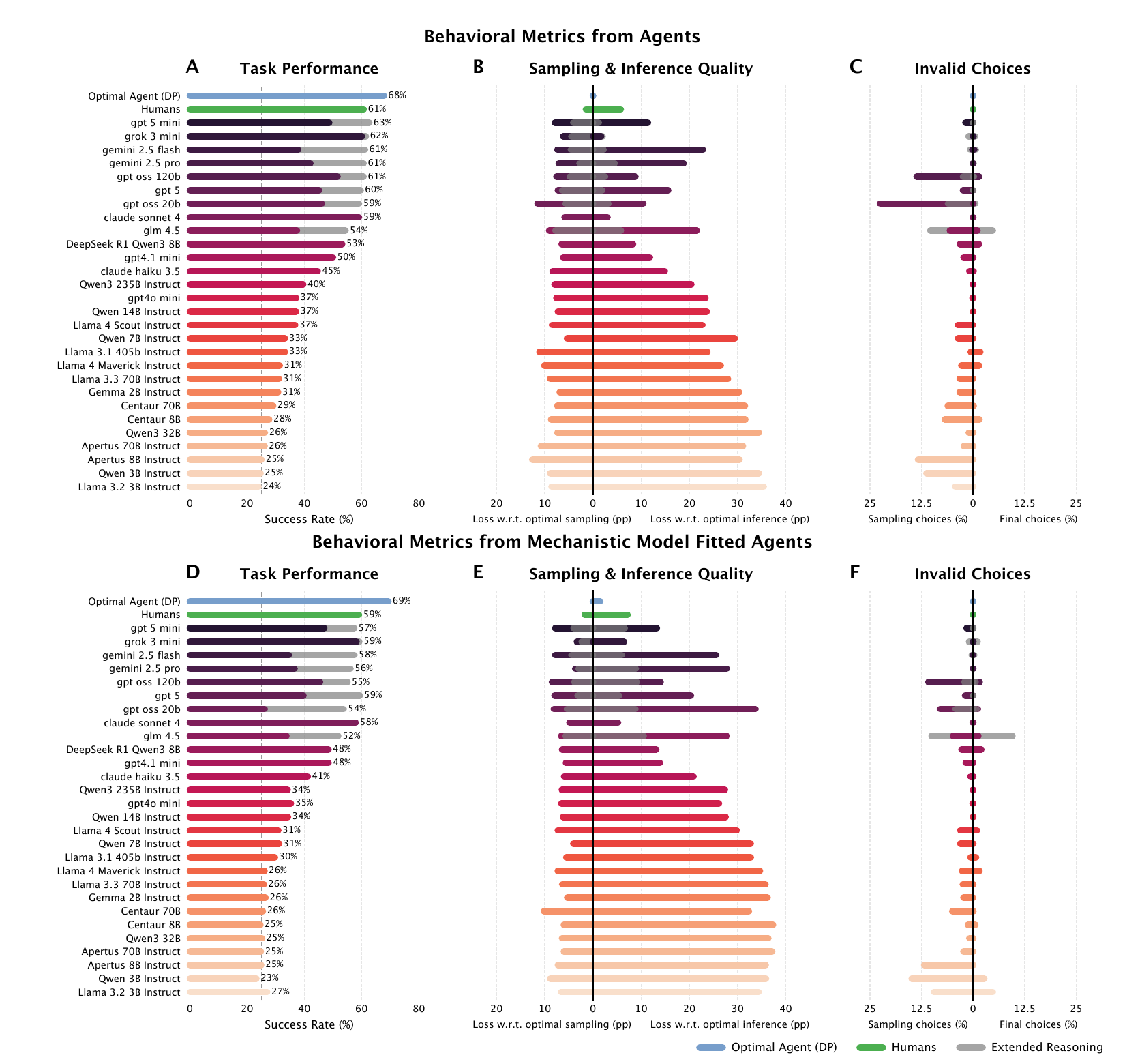}
        \caption{\textbf{Behavioral metrics are recapitulated by the fitted behavioral model.}
        \textbf{A--C}: Empirical behavioral metrics computed directly from human and LLM trajectories. \textbf{A}: Overall task performance (success rate) across agents, as in Fig.~\ref{fig:bench}. \textbf{B}: Decomposition of performance loss into \emph{sampling} versus \emph{inference} deficits, measured by comparing each agent to counterfactual and optimal reference policies (loss w.r.t.\ optimal sampling on the left; loss w.r.t.\ optimal inference on the right). \textbf{C}: Invalid-choice rates during sampling and final report, as in Fig.~\ref{fig:bench}.
        \textbf{D--F}: The same metrics computed from trajectories generated by the \emph{behavioral model} of Fig.~\ref{fig:model_summary} after fitting $(\beta, m, \gamma, \kappa_f, \omega, \theta)$ to each agent. \textbf{D}: Model-predicted success rates closely match empirical performance ordering. \textbf{E}: Model reproduces the sampling--inference loss decomposition, capturing that reasoning primarily reduces inference loss while leaving a residual sampling gap. \textbf{F}: Model reproduces invalid-choice profiles via the occlusion-awareness component. Green denotes pooled humans; grey indicates extended-reasoning conditions.}

        \label{appendix_behaviors}
\end{figure*}

\section{Per model parameter values}
\label{appdx:model_parameters} 
\begin{figure*}[htp]
\centering
\includegraphics[width=\linewidth]{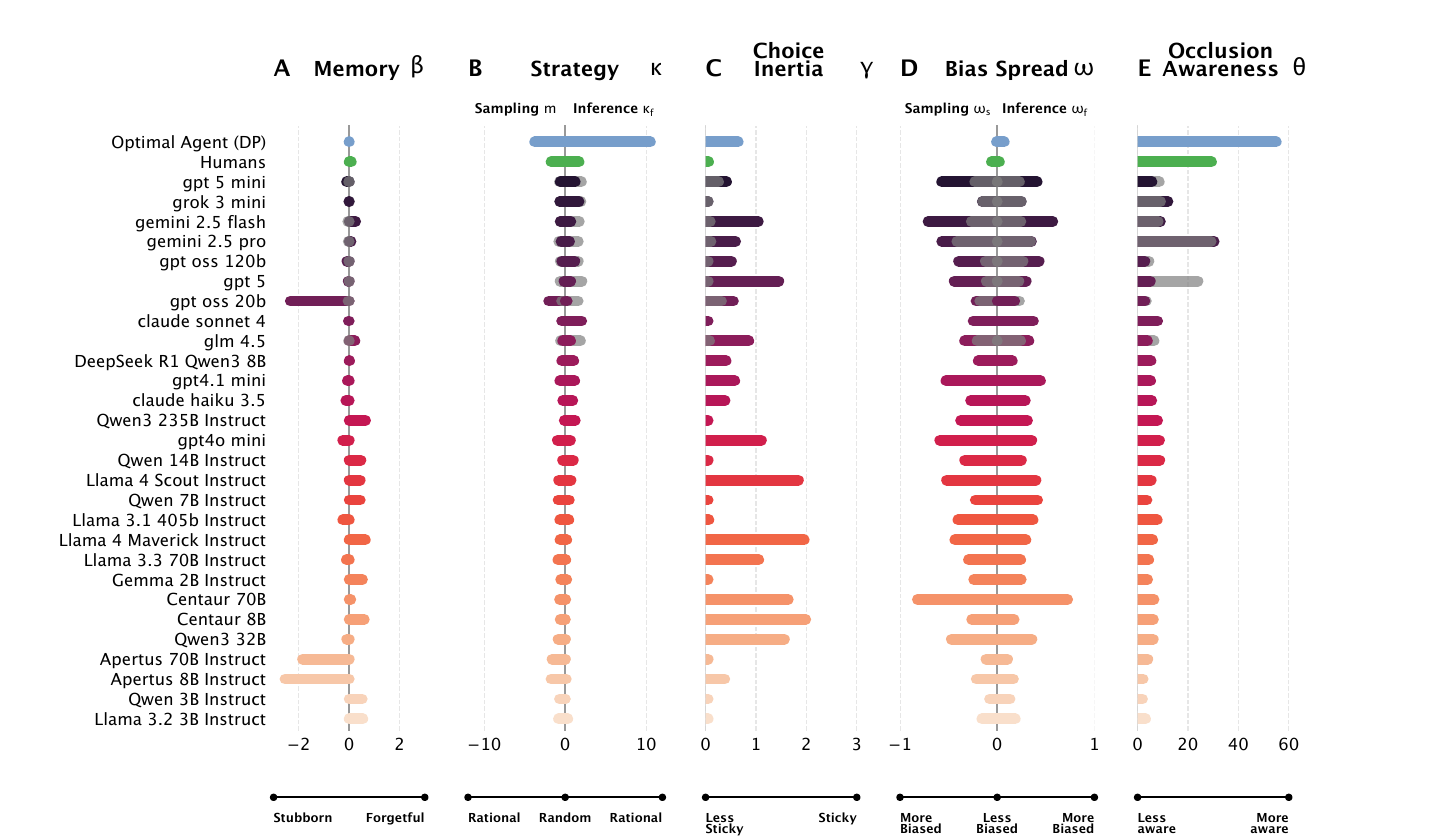}
    \caption{\textbf{Model parameters explain human and LLM behavior.} \textbf{A}: Memory $\beta$ spans \emph{stubborn} ($\beta<0$) to \emph{forgetful} ($\beta>0$) evidence integration. \textbf{B}: Strategy separates sampling gain $m$ in the sampling temporal kernel $k_s(t)$ from final-choice sharpness $\kappa_f$. \textbf{C}: Choice inertia $\gamma$ captures perseveration on the most recent valid sampling choice. \textbf{D}: Bias spread (difference between max and min $\omega_x$ bias vector components) summarizes systematic sampling and inference choice preferences ($\omega_s,\omega_f$). \textbf{E}: Occlusion awareness $\theta$ captures avoidance of invalid or occluded choices.}
\end{figure*}

\paragraph{Parameters underlying sampling and inference loss.} For the 36 LLM conditions, we regressed sampling and inference loss (Tables~\ref{tab:sampling-loss}, \ref{tab:inference-loss}) jointly on the six standardized parameters ($\log(1+\kappa_f)$, $|\beta|$, $m$, $\gamma$, $\log\theta_s$, and bias spread) and decomposed the explained variance into Shapley shares, with 95\% intervals from 1{,}000 bootstrap resamples of conditions. The parameters explain $69\%$ of the variance in sampling loss and $88\%$ in inference loss. Sampling loss is driven by memory $|\beta|$ ($27\%$) and the sampling gain $m$ ($8\%$), both with joint-regression coefficients whose 95\% CI excludes zero; the share of $\kappa_f$ ($20\%$) is not significant once these are included. Inference loss is driven by $\kappa_f$ ($63\%$), with a small contribution of bias spread ($3\%$). The memory effect on sampling loss persists when the three most stubborn conditions ($\beta<-1$) are excluded (partial Spearman $\rho=0.39$ controlling for $\kappa_f$; Fig.~\textbf{\ref{fig:mech_model}J}).

\section{Activation extraction and linear probing}
\label{app:activation_extraction}

\paragraph{Activation extraction.} For each of $50$ games per \textit{reasoning level}, we ran \emph{gpt oss 20b} (24 transformer blocks, $d_{\text{model}}=2880$) through the full task prompt and recorded the residual-stream activations after the MLP of each block at every round (15 sampling rounds plus 1 inference round). Because reasoning traces vary in length and structure across rounds, we summarized each round's activations by averaging over the last 5 tokens of the trace, following standard practice for token-position-robust probing in autoregressive models. The resulting tensor has shape $50 \text{ games} \times 16 \text{ rounds} \times 24 \text{ layers} \times 2 \text{ reasoning levels} \times 2880 \text{ dims}$.

\paragraph{Linear probes.} We fit ridge regression probes that predict the model's choice at each round from its activations, separately for each (layer, reasoning level) pair. Following \citet{alain2016understanding, belinkov2022probing, gurnee2023language}, we use the closed-form ridge solution with the regularization strength $\alpha$ selected by leave-one-out cross-validation over a logarithmic grid $\alpha\in\{10^{-3}, \dots, 10^{6}\}$. We targeted the choice as a 4-way categorical over $\{A,B,C,D\}$ and report held-out $R^2$ averaged across 5 K-fold cross-validation folds (uniform-average over the 4 target dimensions).

\paragraph{Swapped control.} A high probe $R^2$ on activations alone does not tell us whether the decoded choice subspace is specific to the reasoning condition or simply reflects task-level structure shared across conditions (e.g., token identity in the trace). To test for condition-specificity, we trained probes on one condition and evaluated them on the other — denoted \emph{X=high, Y=low (SC)} and \emph{X=low, Y=high (SC)} in Fig.~\textbf{\ref{fig:cognitive_space}D}. SC $R^2$ collapses to near zero or below across all layers, confirming that the choice subspaces are condition-specific.

\section{PCA geometry of layer-1 activations}
\label{app:pca_geometry}

\paragraph{Conditional-average dataset.} A naive PCA over all individual-game activations would mix variance from games, layers, and conditions, obscuring the round-by-round structure of interest. To preserve this structure while suppressing game-level noise, we constructed a conditional-average dataset: for each (round, choice, reasoning level, layer) combination, we averaged activations across the games in which the model made that choice on that round. With $16$ rounds, $4$ choices, $2$ conditions, and $24$ layers, this yields a $16 \times 4 \times 2 \times 24 = 3072$-row matrix in $\mathbb{R}^{2880}$. We restrict our analyses to layer 1 (selected for its high probe $R^2$; Appendix~\textbf{\ref{app:activation_extraction}}) and to a single reasoning level at a time, leaving $64$ rows per PCA fit, but the PCA basis is well-defined at any layer.

\begin{figure*}[htp]
    \centering
    \includegraphics[width=\linewidth]{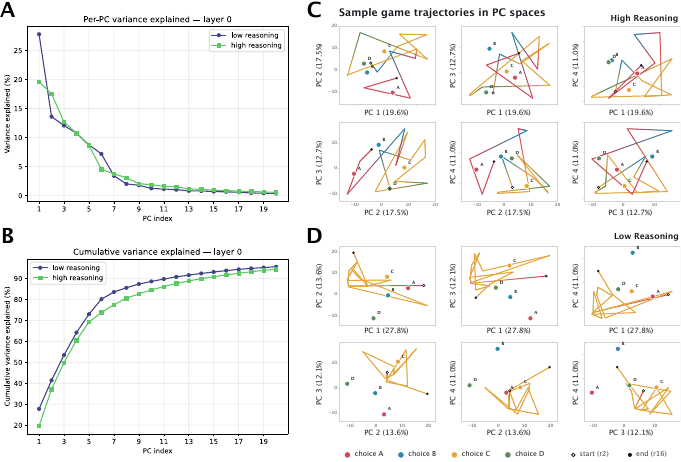}
    \caption{\textbf{PCA geometry of layer-1 activations.}
    \textbf{A:} Per-PC variance explained for the conditional-average dataset at layer 1, shown separately for \textit{low} (purple) and \textit{high} (green) reasoning. PC1 captures $27.8\%$ of the variance under \textit{low reasoning} and $19.6\%$ under \textit{high reasoning}; the flatter spectrum under \textit{high reasoning} indicates richer round-by-round dynamics distributed across more components.
    \textbf{B:} Cumulative variance explained for the same data; the top four PCs together account for $\sim\!64\%$ (\textit{low}) and $\sim\!61\%$ (\textit{high}) of the total variance.
    \textbf{C--D:} Pairwise 2D projections of the example game's layer-1 activations (same game as in Fig.~\textbf{\ref{fig:cognitive_space}}) across the top four PCs, under \textit{high reasoning} (\textbf{C}) and \textit{low reasoning} (\textbf{D}). Per-choice centroids are shown as filled circles colored by choice (A: red, B: blue, C: yellow, D: green); the trajectory's start (round 2) and end (round 16) are marked by open and filled black circles. Across all PC pairs, the \textit{high-reasoning} trajectory consistently migrates from the neighborhood of centroid \emph{C} toward centroid \emph{A} as evidence accumulates, whereas the \textit{low-reasoning} trajectory remains anchored near \emph{C} throughout the game.}
    \label{fig:pairwise_pc}
\end{figure*}

\paragraph{Variance explained.} The top four PCs capture the bulk of the variance in both reasoning conditions (Fig.~\textbf{\ref{fig:pairwise_pc}A--B}): under \textit{low reasoning}, PC1 alone accounts for $27.8\%$ of the variance and the top four PCs together explain $\sim\!64\%$, while under \textit{high reasoning} the variance is more evenly distributed across components, with PC1 capturing $19.6\%$ and the top four PCs reaching $\sim\!61\%$. The flatter spectrum under \textit{high reasoning} is consistent with richer round-by-round dynamics that occupy a higher-dimensional manifold, whereas the steeper spectrum under \textit{low reasoning} reflects trajectories that remain confined to a narrower subspace.

\paragraph{Centroids.} For each (choice, reasoning level) pair, we compute a \emph{centroid} as the mean of its conditional-average vectors across rounds, in the original $2880$-dimensional space. We use these centroids both as projection targets in PCA visualizations (Fig.~\textbf{\ref{fig:cognitive_space}E--F}) and as references for high-dimensional distance computations.

\paragraph{Centroid-distance similarities.} For the example game in Fig.~\textbf{\ref{fig:cognitive_space}}, we compute Euclidean distances $d_k$ from each round's $2880$-dimensional activation to all four choice centroids (within the matching \textit{reasoning} condition), and convert them to a similarity distribution via $p_k \propto \exp(-d_k)$. This places the resulting 4-vector on a simplex enabling direct visual comparison between the model's latent variables and a quantity computed entirely from raw activations (Fig.~\textbf{\ref{fig:cognitive_space}B--C}, bottom row). For visualization, we additionally project per-round activations into the layer-1 PCA basis and color the trajectory by each round's \emph{closest centroid}, i.e.\ the $\arg\min$ over the same distances $d_k$ (Fig.~\textbf{\ref{fig:cognitive_space}E--F}).

\paragraph{Pairwise PC views.} Three-dimensional projections can occasionally distort relationships between centroids that are well-separated in higher dimensions. As a robustness check, we provide pairwise 2D projections across the top four PCs (Fig.~\textbf{\ref{fig:pairwise_pc}C} for \textit{high reasoning} and Fig.~\textbf{\ref{fig:pairwise_pc}D} for \textit{low reasoning}); the qualitative pattern: \textit{high-reasoning} trajectories migrating from \emph{C} to \emph{A}, \textit{low-reasoning} trajectories remaining near \emph{C}, is consistent across PC pairs.

\section{Model Formulation Details}
\label{appdx:model_formulation}
We define the behavior model as a structured set of departures from an ideal Bayesian observer. Agents may integrate evidence imperfectly, show systematic choice preferences, ignore occlusions and make invalid choices, or translate beliefs into choices with either MAP-like exploitation or near-random exploration. The goal is to characterize each fitted agent by how evidence is accumulated over time and how the resulting beliefs are converted into sampling and final choices.

\subsection{Task likelihood and Bayesian reference}

We consider independent games indexed by $g$. Game $g$ contains $T_g$\footnote{note that in the main text we drop the game index $g$ for simplicity.} sampling rounds followed by one final inference choice. There are $K$ possible cues, with $K=4$ in the task. Each game has a latent biased cue $z_g\in\{1,\dots,K\}$ with prior $\bm p_0\in\Delta^K$, $(\bm p_0)_k=\Pr(z_g=k)$. At sampling round $t$, the agent chooses $a_{g,t}\in\{1,\dots,K\}$ and, when the response yields evidence, observes $x_{g,t}\in\{0,1\}$, where $1$ denotes RED and $0$ denotes GREEN.

The ideal observer assumes the shared-rate Bernoulli emission model
\begin{equation}
p(x_{g,t}\mid a_{g,t}=i,z_g=k)
=
\begin{cases}
\Ber(x_{g,t};\alpha_B), & k=i,\\
\Ber(x_{g,t};\alpha_U), & k\neq i,
\end{cases}
\qquad
0<\alpha_U<\alpha_B<1.
\label{eq:compact_bern_shared_rates}
\end{equation}
Conditional on $z_g$ and the choices, observations are independent across rounds:
\begin{equation}
p(x_{g,1:t}\mid a_{g,1:t},z_g=k)
=
\prod_{u=1}^{t}p(x_{g,u}\mid a_{g,u},z_g=k).
\label{eq:compact_fact_lik}
\end{equation}

For a single game, suppress $g$ when clear and define the likelihood vector
\begin{equation}
(\bm L_t)_k:=p(x_t\mid a_t,z=k).
\label{eq:compact_likelihood_vector}
\end{equation}
The Bayesian posterior is updated by elementwise multiplication and normalization,
\begin{equation}
\bm p_t
=
\frac{\bm p_{t-1}\odot\bm L_t}
{\one^\top(\bm p_{t-1}\odot\bm L_t)}.
\label{eq:compact_bayes_vector}
\end{equation}
Equivalently, with unnormalized log-posterior
\begin{equation}
\bm\ell_t
:=
\log\bm p_0+\sum_{u=1}^{t}\log\bm L_u,
\qquad
\bm\ell_t=\bm\ell_{t-1}+\log\bm L_t,
\label{eq:compact_ell_def}
\end{equation}
and softmax $\sigma(\bm v):=\exp(\bm v)/[\one^\top\exp(\bm v)]$, the posterior is
\begin{equation}
\bm p_t=\sigma(\bm\ell_t).
\label{eq:compact_posterior_softmax}
\end{equation}
This follows because $\exp(\bm\ell_t)$ is the vector of Bayes numerators. The softmax also satisfies $\sigma(\bm v+\alpha\one)=\sigma(\bm v)$ for any scalar $\alpha$, so only relative log-beliefs affect the posterior.

\subsection{Centered evidence and gamma memory}

The shared-rate model implies that, on each round, only the chosen cue receives discriminative evidence relative to the shared mismatch likelihood. Define
\begin{equation}
L_t^+ := p(x_t\mid a_t,z=a_t),
\qquad
L_t^- := p(x_t\mid a_t,z\neq a_t),
\label{eq:compact_lplus_lminus}
\end{equation}
and the single-step log-likelihood ratio
\begin{equation}
\delta_t
:=
\log\frac{L_t^+}{L_t^-}
=
x_t r+(1-x_t)g,
\qquad
r:=\log\frac{\alpha_B}{\alpha_U},
\quad
g:=\log\frac{1-\alpha_B}{1-\alpha_U}.
\label{eq:compact_delta_def}
\end{equation}
If $e_{a_t}\in\R^K$ is the one-hot vector for the sampled cue, then
\begin{equation}
\log\bm L_t=\log(L_t^-)\one+\delta_t e_{a_t}.
\label{eq:compact_logL_decomp}
\end{equation}

Because softmax is invariant to common shifts, we work in centered log-belief space. Let
\begin{equation}
C:=I-\frac{1}{K}\one\one^\top,
\qquad
C\one=\bm 0,
\qquad
C^2=C.
\label{eq:compact_C_def}
\end{equation}
The Bayesian centered evidence increment is
\begin{equation}
\Delta\bm h_t
:=
C\log\bm L_t
=
\delta_t\left(e_{a_t}-\frac{1}{K}\one\right).
\label{eq:compact_delta_h}
\end{equation}
Thus the ideal centered integrator accumulates the same evidence as Bayes, but removes the unidentifiable global baseline. When all increments are accumulated equally, $\sigma(\sum_{u=1}^{t}\Delta\bm h_u)=\bm p_t$ up to the prior-centered initial state.

The fitted model replaces exact accumulation with a temporal memory kernel. This kernel captures order effects in belief updating: older evidence can be discounted, preserved, or overweighted, matching recency and primacy distortions studied in leaky-accumulator and belief-adjustment accounts \citep{usher2001time,hogarth1992order}. We use an exponential form because it makes recency and primacy deviations symmetric in log-weight space. For equal $|\beta|$, the log weights have matched slopes but tilt toward the current round when $\beta>0$ and toward the start of the sequence when $\beta<0$.
\begin{equation}
\bm h_{g,t}
:=
\sum_{u=1}^{t}w_{t,u}(\beta)\Delta\bm h_{g,u},
\label{eq:compact_h_kernel}
\end{equation}
with
\begin{equation}
w_{t,u}(\beta)
=
\begin{cases}
\exp[-\beta(t-u)], & \beta>0,\\
1, & \beta=0,\\
\exp[\beta(u-1)], & \beta<0.
\end{cases}
\label{eq:compact_h_kernel_weights}
\end{equation}
Here $\beta=0$ recovers exact accumulation, $\beta>0$ produces forgetful recency weighting, and $\beta<0$ produces stubborn primacy weighting.

\subsection{Internal posterior and policy}

Following bounded-rationality formulations with information-processing costs \citep{ortega2013thermodynamics}, choices are generated from softmax-sharpened internal posteriors. For the selected gamma fit, sampling sharpness depends on how many valid observations are available before the choice:
\begin{equation}
n^-_{g,t}:=1+\sum_{u<t}\mathbf 1\{a_{g,u}\ \mathrm{valid}\},
\qquad
\kappa_s(g,t):=m\,\frac{n^-_{g,t}}{T_g}.
\label{eq:compact_observed_count_schedule}
\end{equation}
This schedule makes sampling initially exploratory and increasingly belief-guided as usable evidence accumulates, consistent with exploration--exploitation accounts \citep{daw2011trial,gershman2019uncertainty}.

Let $\bm h^-_{g,t}$ be the pre-choice memory state at sampling round $t$. The sharpened sampling and final-choice posteriors are
\begin{equation}
\bm q^s_{g,t}:=\sigma\!\left(\kappa_s(g,t)\bm h^-_{g,t}\right),
\qquad
\bm q^f_g:=\sigma(\kappa_f\bm h_{g,T_g}).
\label{eq:compact_internal_softmax_kh}
\end{equation}
Here $m$ controls how quickly sampling becomes evidence-guided, while $\kappa_f$ controls final-choice sharpness.

During sampling, not all cues are available. Let $\bm v_{g,t}\in\{0,1\}^K$ be the availability vector and let $\bm i_{g,t}\in\{0,1\}^K$ be one-hot for the most recent valid sampling choice, with $\bm i_{g,t}=\bm 0$ if no valid choice has occurred. It is clearest to write the cue-response policy in log-score space, where the belief, bias, inertia, and occlusion terms add; $\log\bm\omega_s$ and $\log\bm\omega_f$ are elementwise logs of positive bias vectors:
\begin{equation}
\bm\eta^s_{g,t}
:=
\kappa_s(g,t)\bm h^-_{g,t}
+\log\bm\omega_s
+\gamma\bm i_{g,t}
-\theta_s(\one-\bm v_{g,t}),
\qquad
\bm\pi^s_{g,t}:=\sigma(\bm\eta^s_{g,t}),
\label{eq:compact_sampling_logscore_policy}
\end{equation}
\begin{equation}
\bm\eta^f_g
:=
\kappa_f\bm h_{g,T_g}
+\log\bm\omega_f,
\qquad
\bm\pi^f_g:=\sigma(\bm\eta^f_g).
\label{eq:compact_final_logscore_policy}
\end{equation}

Larger $\theta_s$ means stronger avoidance of occluded sampling actions, and larger $\gamma$ means stronger repeat-choice inertia. Equivalently, for any two cues $j,k$, the sampling and final log odds are
\begin{align}
\log\frac{(\bm\pi^s_{g,t})_j}{(\bm\pi^s_{g,t})_k}
&=
\left(b^s_{g,t,j}-b^s_{g,t,k}\right)
+\log\frac{(\bm\omega_s)_j}{(\bm\omega_s)_k}
+\gamma\left(i_{g,t,j}-i_{g,t,k}\right)
\nonumber\\
&\hspace{1.5cm}
-\theta_s\left[(1-v_{g,t,j})-(1-v_{g,t,k})\right],
\label{eq:compact_sampling_logodds_policy}
\\
\log\frac{(\bm\pi^f_g)_j}{(\bm\pi^f_g)_k}
&=
\left(b^f_{g,j}-b^f_{g,k}\right)
+\log\frac{(\bm\omega_f)_j}{(\bm\omega_f)_k}.
\label{eq:compact_final_logodds_policy}
\end{align}
Here $\bm b^s_{g,t}:=\kappa_s(g,t)\bm h^-_{g,t}$ and $\bm b^f_g:=\kappa_f\bm h_{g,T_g}$ are the sampling and final belief log-scores.

The same policy can be written in multiplicative form by exponentiating these log scores. Define
\begin{equation}
\bm r_{g,t}:=\exp(\gamma\bm i_{g,t}),
\qquad
\bm o^s_{g,t}:=\exp[-\theta_s(\one-\bm v_{g,t})].
\label{eq:compact_hadamard_factors}
\end{equation}
Then Eq.~\eqref{eq:compact_sampling_logscore_policy}--\eqref{eq:compact_final_logscore_policy} are equivalently
\begin{equation}
\bm u^s_{g,t}
=
\exp(\bm b^s_{g,t})
\odot\bm\omega_s
\odot\bm r_{g,t}
\odot\bm o^s_{g,t},
\qquad
\bm\pi^s_{g,t}
=
\frac{\bm u^s_{g,t}}{\one^\top\bm u^s_{g,t}},
\label{eq:compact_sampling_hadamard_policy}
\end{equation}
\begin{equation}
\bm u^f_g
=
\exp(\bm b^f_g)
\odot\bm\omega_f,
\qquad
\bm\pi^f_g
=
\frac{\bm u^f_g}{\one^\top\bm u^f_g}.
\label{eq:compact_final_hadamard_policy}
\end{equation}
The log-score form makes the additive contributions transparent, while the multiplicative form shows the equivalent implementation: exponentiated belief scores are combined with bias, inertia, and occlusion factors before normalization.

\subsection{Model loss function}

For each agent, the dataset consists of $G$ games: a balanced subset of $100$ games for each trial length $T\in\{2,\dots,15\}$ ($G=1{,}400$; $1{,}200$ for \emph{apertus 70b}, which has no games of length $11$ or $12$), except for humans, whose $4{,}600$ games are all used. Game $g$ provides sampling choices $\{a_{g,t}\}_{t=1}^{T_g}$, observed outcomes $\{x_{g,t}\}_{t=1}^{T_g}$, sampling occlusion sets $\{\mathcal A_{g,t}\}_{t=1}^{T_g}$, and a final inference choice $f_g$. Given parameters
\begin{equation}
\Theta := (\bm\omega_s,\bm\omega_f,\beta,m,\gamma,\kappa_f,\theta_s,\theta_f),
\label{eq:compact_theta}
\end{equation}
we unroll the memory trajectory deterministically from the observed history, then compute the policies above. In the implementation, $\theta_s$ is the fitted sampling occlusion penalty previously named $\theta_{\mathrm{occ}}$, and the fitted invalid-response probability is represented through
\begin{equation}
\theta_f=\log\frac{1-p_{\mathrm{abstain}}}{p_{\mathrm{abstain}}},
\qquad
P(\mathrm{invalid/missing})=\sigma(-\theta_f).
\label{eq:compact_theta_f_map}
\end{equation}

For any observed response $y$ and cue policy $\bm\pi$, define
\begin{equation}
P_{\Theta}(y\mid\bm\pi,\theta_f)
=
\begin{cases}
\sigma(\theta_f)(\bm\pi)_y, & y\in\{1,\dots,K\},\\
\sigma(-\theta_f), & y\ \mathrm{invalid/missing}.
\end{cases}
\label{eq:compact_response_probability}
\end{equation}
The negative cue-response log-likelihood is
\begin{equation}
\mathcal N(\Theta)
=
-\sum_{g=1}^{G}
\left[
\sum_{t=1}^{T_g}
\log P_{\Theta}(a_{g,t}\mid\bm\pi^s_{g,t},\theta_f)
+
\log P_{\Theta}(f_g\mid\bm\pi^f_g,\theta_f)
\right].
\label{eq:compact_loglik}
\end{equation}

The minimized objective is
\begin{equation}
\mathcal J(\Theta)
=
\mathcal N(\Theta)
+\lambda_{\omega}\,\mathrm{KL}\!\left(\bm\omega_s\,\|\,\tfrac{1}{K}\one\right)
+\lambda_{\gamma}\,\gamma^2 .
\label{eq:compact_regularized_nll}
\end{equation}

\begin{table}[t]
\centering
\footnotesize
\caption{Selected gamma-fit objective weights and parameter bounds. Alignment losses are ordered as written in the text.}
\label{tab:compact_gamma_fit_objective_settings}
\begin{tabular}{lll}
\toprule
Setting & Symbol & Selected value \\
\midrule
Choice likelihood weight & $\lambda_{\mathrm{choice}}$ & $1.0$ \\
Sampling-bias regularizer & $\lambda_{\omega}$ & $0.005$ \\
Inertia regularizer & $\lambda_{\gamma}$ & $0.005$ \\
\midrule
Memory bound & $\beta$ & $[-3,3]$ \\
Sampling-gain bound & $m$ & $[0.05,7]$ \\
Inertia bound & $\gamma$ & $[0.05,3]$ \\
Sampling occlusion bound & $\theta_s$ & $[0,60]$ \\
Final sharpness bound & $\kappa_f$ & $(0,7)$ \\
Invalid-response logit bound & $\theta_f$ & $[0,60]$ \\
\bottomrule
\end{tabular}
\end{table}

\subsection{Fitting Results}
\label{appdx:fitting_results}

\begin{figure}[htp]
        \centering
        \includegraphics[width=0.8\textwidth]{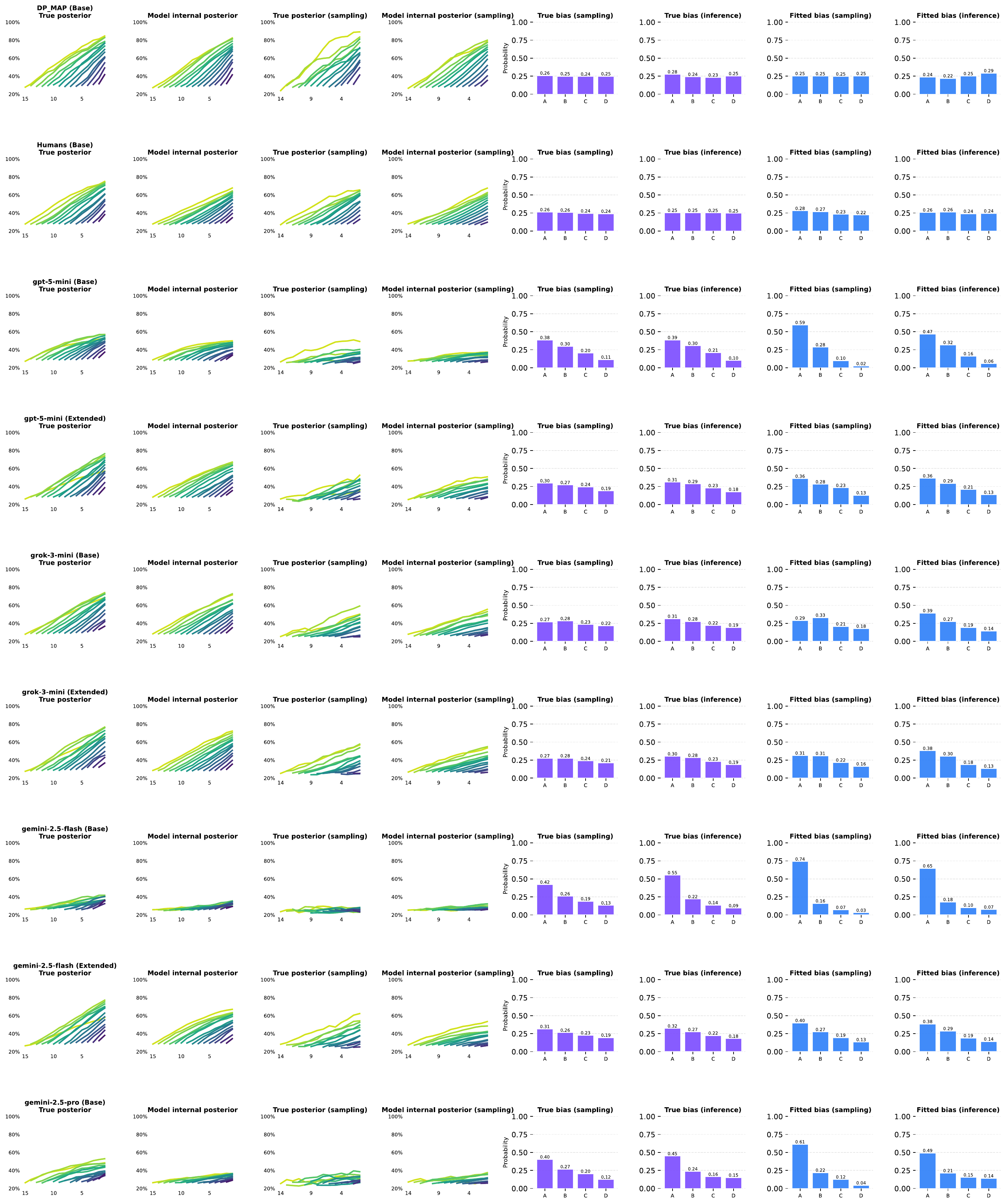}
        \caption{\textbf{Posterior evolution by rounds and bias across humans and language models.} In each row the first two panels show the true and model-generated posterior evolution for the final choice (\textit{inference}), followed by the true and model-generated sampling-choice posterior (\textit{sampling}). The remaining panels show the true bias for sampling and inference and their corresponding fitted bias weights.}
        \label{appendix:postbias1}
\end{figure}

\begin{figure}[htp]
        \centering
        \includegraphics[width=0.94\textwidth]{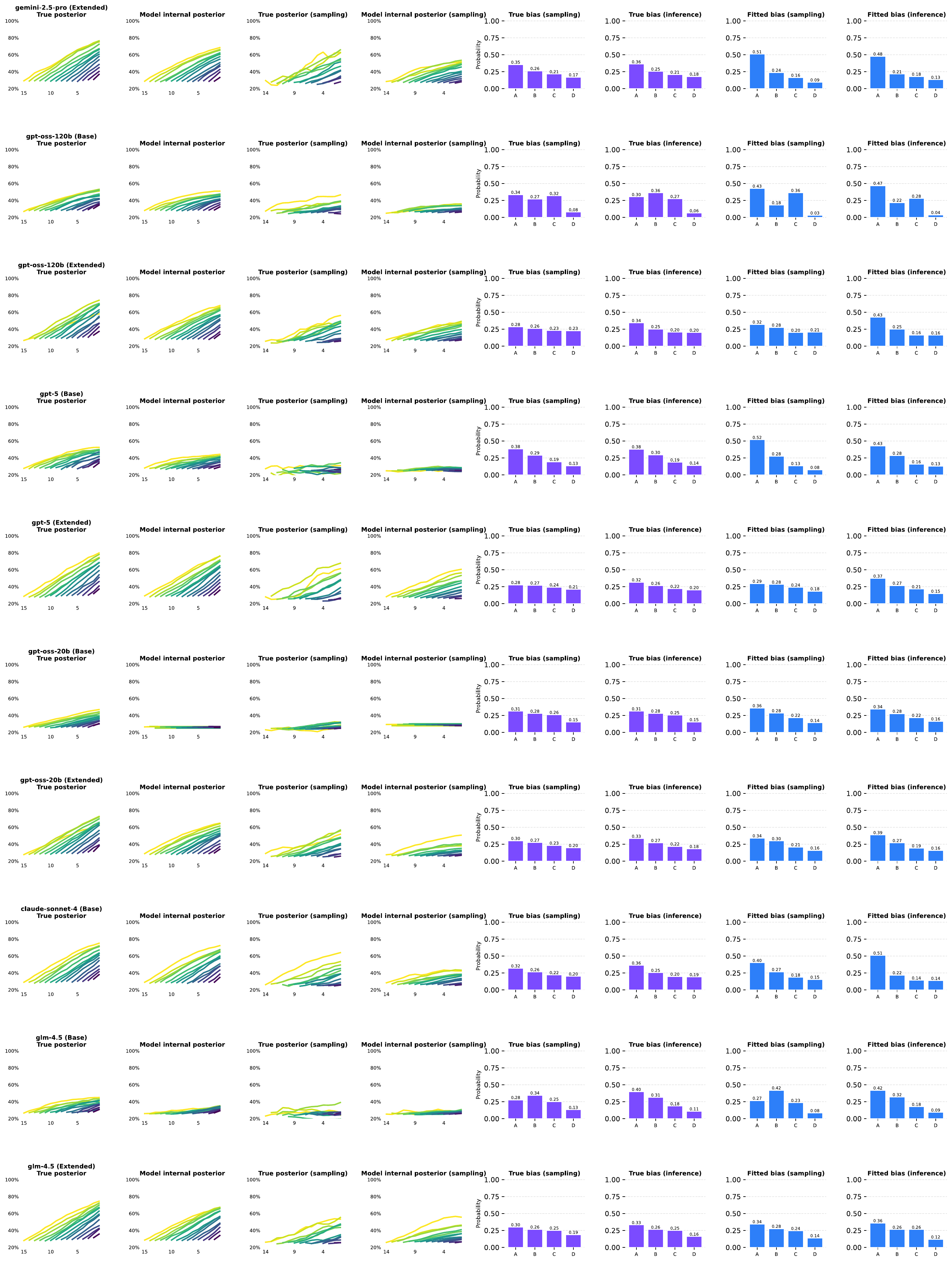}
        \caption{\textbf{Posterior evolution by rounds and bias across humans and language models.} In each row the first two panels show the true and model-generated posterior evolution for the final choice (\textit{inference}), followed by the true and model-generated sampling-choice posterior (\textit{sampling}). The remaining panels show the true bias for sampling and inference and their corresponding fitted bias weights.}
        \label{appendix:postbias2}
\end{figure}

\begin{figure}[htp]
        \centering
        \includegraphics[width=0.94\textwidth]{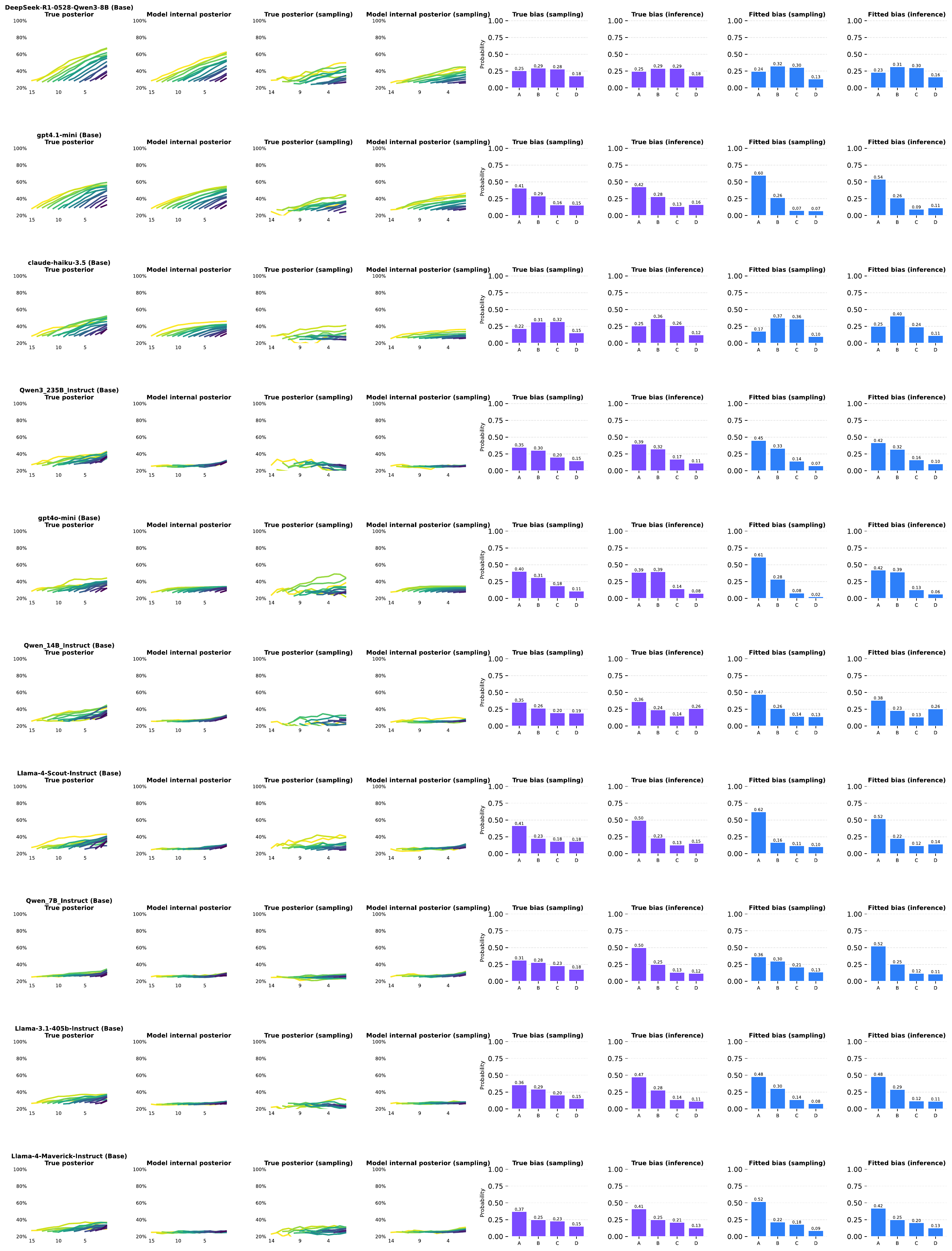}
        \caption{\textbf{Posterior evolution by rounds and bias across humans and language models.} In each row the first two panels show the true and model-generated posterior evolution for the final choice (\textit{inference}), followed by the true and model-generated sampling-choice posterior (\textit{sampling}). The remaining panels show the true bias for sampling and inference and their corresponding fitted bias weights.}
        \label{appendix:postbias3}
\end{figure}

\begin{figure}[htp]
        \centering
        \includegraphics[width=0.94\textwidth]{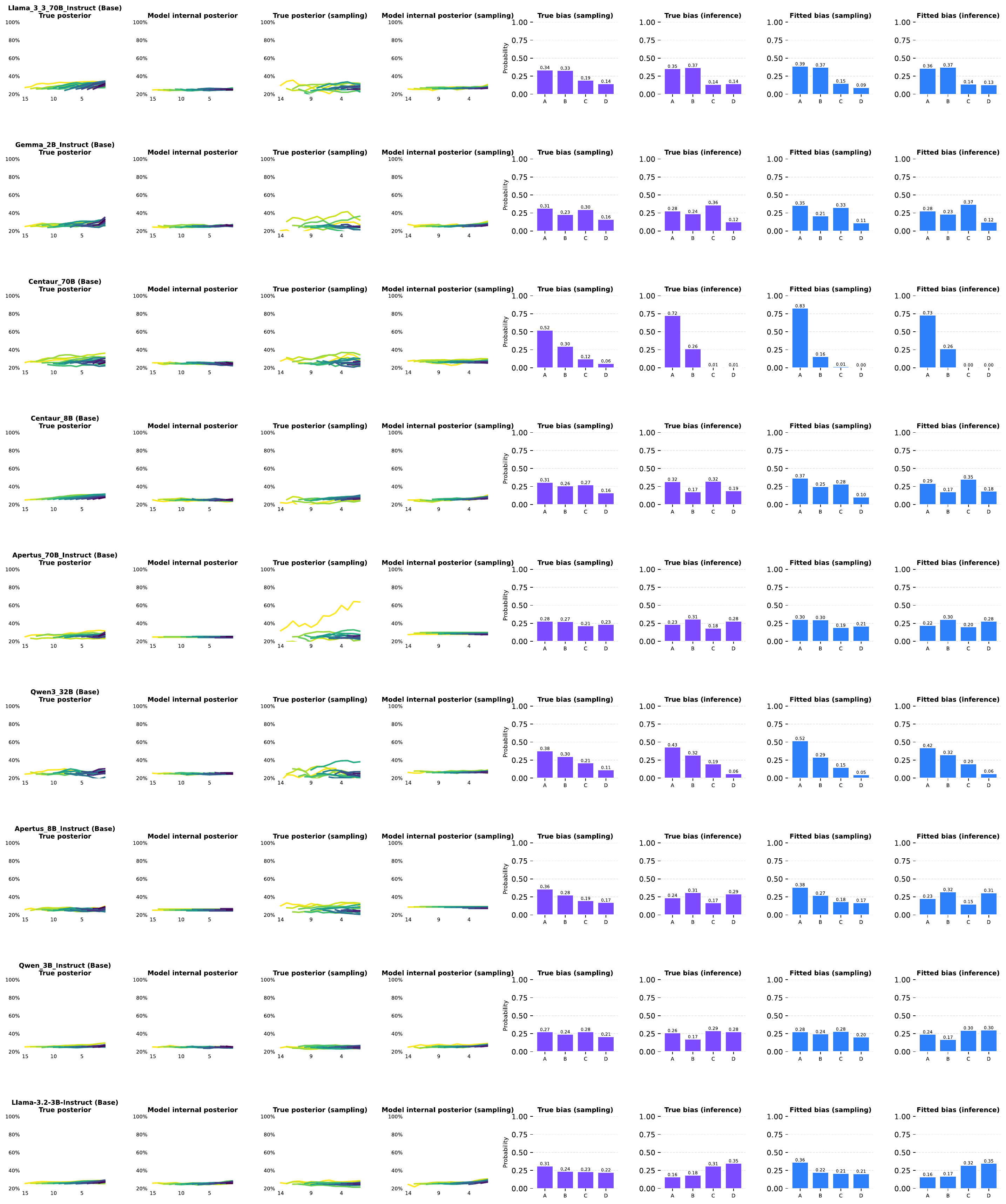}
        \caption{\textbf{Posterior evolution by rounds and bias across humans and language models.} In each row the first two panels show the true and model-generated posterior evolution for the final choice (\textit{inference}), followed by the true and model-generated sampling-choice posterior (\textit{sampling}). The remaining panels show the true bias for sampling and inference and their corresponding fitted bias weights.}
        \label{appendix:postbias4}
\end{figure}

For each agent (humans and each LLM) and reasoning mode (\textit{Base}/\textit{Extended}), Appendix Figs.~\ref{appendix:postbias1}--\ref{appendix:postbias4} visualize (i) posterior trajectories during \textit{inference} and \textit{sampling}, and (ii) choice biases alongside the fitted bias weights $\{\bm\omega_s,\bm\omega_f\}$. The same quantities are summarized as heatmaps in Appendix Fig.~\ref{appendix_posterior_bias_heatmap}. Full fitted parameter values are reported in Table~\ref{appdx:model_parameter_table} (corresponding to Fig.~\ref{fig:mech_model} in the main text).

\paragraph{Posterior curves.} All posterior panels compress the full $K$-dimensional belief vector into a single scalar by reading out the posterior mass assigned to the choice that was actually taken. Concretely, let $\bm p_{g,t}\in\Delta^K$ denote the (normative) Bayesian posterior after $t$ samples in game $g$. Then for \emph{inference} panels (columns 1--2), the plotted quantity is the posterior probability of the \emph{final chosen option}, evaluated as evidence accumulates across rounds. For \emph{sampling} panels (columns 3--4), the plotted quantity is the posterior probability of the \emph{sampling action chosen on round $t$}, evaluated \emph{just before} the sample is taken on that round (so that it reflects the beliefs that informed the sampling decision).

Curves are grouped by trial length $N\in\{2,\dots,15\}$. For a fixed $N$, we compute the scalar trajectory above for every game of length $N$ and then average across games, producing one mean trajectory per $N$ (shown as a family of curves, ordered by $N$). Thus, each line summarizes the \emph{average posterior dynamics conditional on game length} for \emph{inference} and \emph{sampling}.

\textbf{True posteriors} are computed from the observed trajectories of each agent using the normative Bayesian update (Eq.~\ref{eq:normative_update}). \textbf{Fitted posterior readouts} are generated by unrolling the fitted gamma model and reading out the corresponding sharpened internal posteriors.

\paragraph{Bias panels.}
The bias panels compare empirical choice frequencies against the fitted bias weights. Empirical bars report how often each option is chosen (separately for sampling and inference), while the fitted bars show the learned bias vectors $\bm\omega_s$ and $\bm\omega_f$ (or their summary statistic, depending on the plot), which capture systematic preferences not explained by the posterior.

\begin{figure*}[htp]
        \centering
        \includegraphics[width=1.0\textwidth]{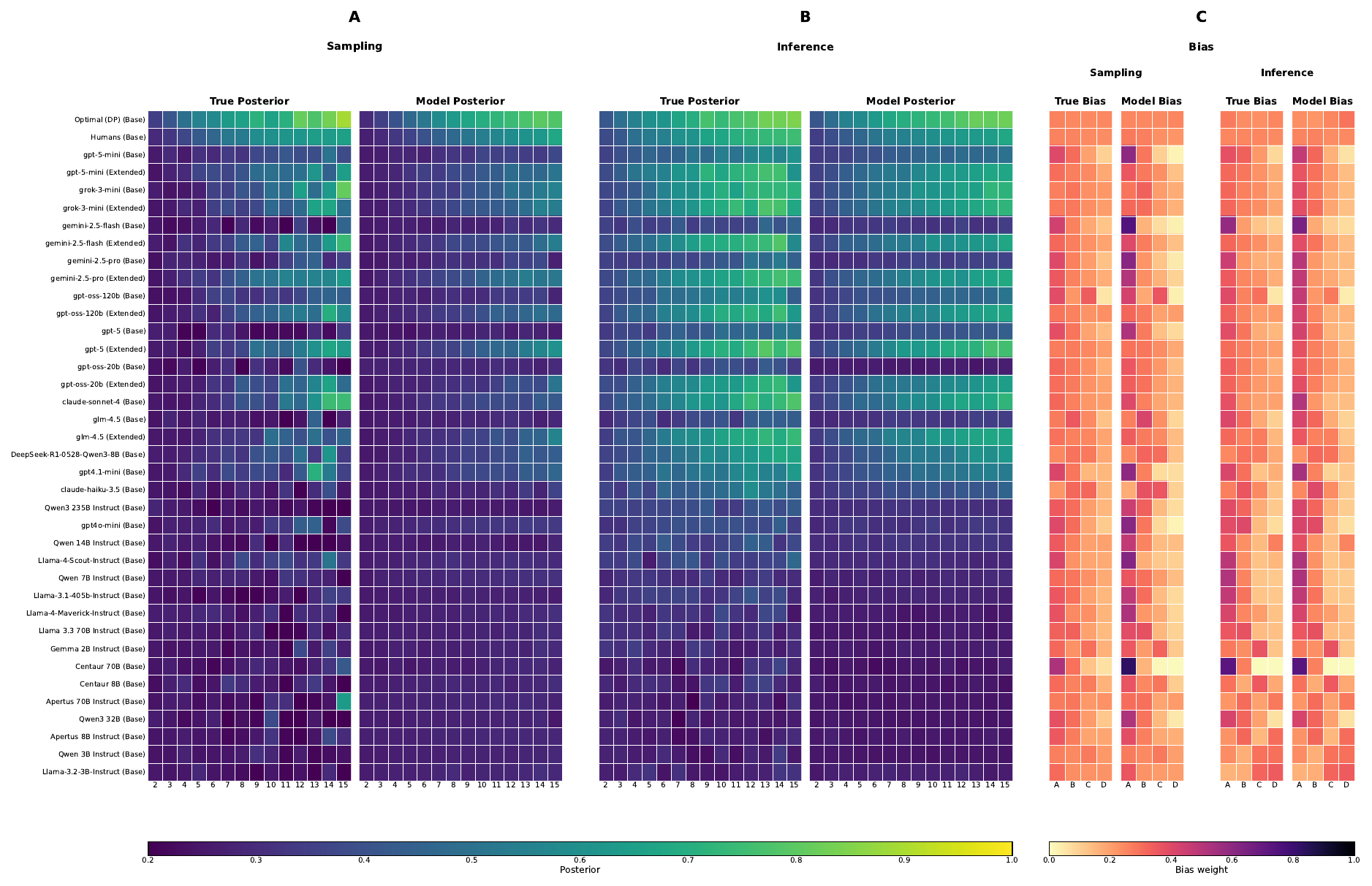}
        \caption{\textbf{Posterior and bias structure is captured by the mechanistic model.}
        For each agent (rows), we compute phase-specific summaries during \textbf{sampling} (left block) and \textbf{inference} (middle block), and compare empirical estimates (\textbf{True}) to quantities generated by the fitted mechanistic model (\textbf{Model}). \textbf{Sampling}: heatmaps show the agent's \textbf{True Posterior} signal (normative posterior mass assigned to the agent's eventual choice under the game history) alongside the corresponding \textbf{Model Posterior} obtained by unrolling the fitted belief dynamics. \textbf{Inference}: analogous comparison for the final inference phase. \textbf{Bias}: right block contrasts \textbf{True Bias} (empirical action preference beyond belief) with \textbf{Model Bias} recovered from the multiplicative bias parameters. Across agents, the model reproduces both the magnitude and qualitative structure of posterior dynamics and systematic biases, supporting the claim that $(\beta,m,\kappa_f,\omega,\theta_s,\theta_f,\gamma)$ provide a sufficient explanation of the behavioral signatures.}
        \label{appendix_posterior_bias_heatmap}
\end{figure*}

\subsection{Statistical significance of behavioral model fits}
\label{app:behav_stats}
\nopagebreak[4]
\vspace{0.25\baselineskip}
\begingroup
\scriptsize
\setlength{\tabcolsep}{8pt}
\setlength{\LTleft}{\fill}
\setlength{\LTright}{\fill}
\setlength{\LTpre}{0pt}
\setlength{\LTpost}{0pt}
\begin{longtable}{>{\raggedright\arraybackslash}p{3.8cm} l r r r r r}
\caption{Per held-out game, improvement is $\delta_i = \log p(y_i\mid M) - \log p(y_i\mid B)$, where $M$ is the fitted model and $B$ is an agent-specific frequency baseline. Reported $\Delta$eLPD/game is $\bar{\delta}=\frac{1}{N}\sum_{i=1}^{N}\delta_i$ across out-of-sample games (repeated 5-fold CV; repeated predictions per game averaged first). \textbf{CI low/high}: bootstrap confidence interval for $\bar{\delta}$. \textbf{p-value}: sign-flip permutation test for $H_0:\bar{\delta}\le 0$ vs $H_1:\bar{\delta}>0$ (multiple-testing adjusted). \textbf{FI}: fraction of held-out games with $\delta_i>0$.}\label{tab:behavioral-model-fits}\\
\toprule
Model & Reasoning & $\Delta$eLPD/game & CI low & CI high & p-value & FI \\
\midrule
\endfirsthead
\caption[]{Statistical significance of behavioral model fits (continued).}\\
\toprule
Model & Reasoning & $\Delta$eLPD/game & CI low & CI high & p-value & FI \\
\midrule
\endhead
Optimal agent (DP) & -- & 8.953 & 8.707 & 9.196 & $<$0.001 & 99.6\% \\
Humans & -- & 6.518 & 6.401 & 6.636 & $<$0.001 & 98.1\% \\
\midrule
\multicolumn{7}{c}{Base models} \\
\midrule
gpt-5-mini & Base & 6.962 & 6.756 & 7.160 & $<$0.001 & 98.6\% \\
grok-3-mini & Base & 7.354 & 7.159 & 7.548 & $<$0.001 & 98.4\% \\
gemini-2.5-flash & Base & 7.716 & 7.492 & 7.936 & $<$0.001 & 99.5\% \\
gemini-2.5-pro & Base & 7.239 & 7.034 & 7.442 & $<$0.001 & 99.8\% \\
gpt-oss-120b & Base & 3.807 & 3.664 & 3.945 & $<$0.001 & 94.7\% \\
gpt-5 & Base & 6.065 & 5.879 & 6.255 & $<$0.001 & 98.6\% \\
gpt-oss-20b & Base & 3.780 & 3.580 & 3.979 & $<$0.001 & 88.9\% \\
claude-sonnet-4 & Base & 7.654 & 7.451 & 7.854 & $<$0.001 & 98.9\% \\
glm-4.5 & Base & 5.911 & 5.714 & 6.106 & $<$0.001 & 97.3\% \\
DeepSeek-R1-0528-Qwen3-8B & Base & 6.663 & 6.457 & 6.856 & $<$0.001 & 98.5\% \\
gpt4.1-mini & Base & 6.526 & 6.333 & 6.728 & $<$0.001 & 98.3\% \\
claude-haiku-3.5 & Base & 7.016 & 6.808 & 7.223 & $<$0.001 & 98.8\% \\
Qwen3\_235B\_Instruct & Base & 4.574 & 4.030 & 5.090 & $<$0.001 & 88.8\% \\
gpt4o-mini & Base & 7.678 & 7.449 & 7.900 & $<$0.001 & 99.1\% \\
Qwen\_14B\_Instruct & Base & 7.124 & 6.931 & 7.318 & $<$0.001 & 99.2\% \\
Llama-4-Scout-Instruct & Base & 6.993 & 6.770 & 7.223 & $<$0.001 & 97.6\% \\
Qwen\_7B\_Instruct & Base & 5.333 & 5.134 & 5.527 & $<$0.001 & 94.2\% \\
Llama-3.1-405b-Instruct & Base & 7.288 & 7.083 & 7.493 & $<$0.001 & 99.3\% \\
Llama-4-Maverick-Instruct & Base & 6.133 & 5.931 & 6.337 & $<$0.001 & 96.8\% \\
Llama\_3\_3\_70B\_Instruct & Base & 5.676 & 5.489 & 5.871 & $<$0.001 & 96.5\% \\
Gemma\_2B\_Instruct & Base & 5.402 & 5.195 & 5.612 & $<$0.001 & 95.1\% \\
Centaur\_70B & Base & 6.149 & 5.852 & 6.444 & $<$0.001 & 92.2\% \\
Centaur\_8B & Base & 6.249 & 6.054 & 6.438 & $<$0.001 & 98.0\% \\
Apertus\_70B\_Instruct & Base & 5.207 & 5.030 & 5.384 & $<$0.001 & 99.0\% \\
Qwen3\_32B & Base & 6.569 & 6.363 & 6.779 & $<$0.001 & 97.6\% \\
Apertus\_8B\_Instruct & Base & 2.850 & 2.690 & 3.008 & $<$0.001 & 86.6\% \\
Qwen\_3B\_Instruct & Base & 2.323 & 2.140 & 2.505 & $<$0.001 & 80.1\% \\
Llama-3.2-3B-Instruct & Base & 4.860 & 4.643 & 5.065 & $<$0.001 & 92.7\% \\
\midrule
\multicolumn{7}{c}{Extended reasoning models} \\
\midrule
gpt-5-mini & Extended & 7.805 & 7.601 & 8.009 & $<$0.001 & 99.3\% \\
grok-3-mini & Extended & 7.239 & 7.038 & 7.439 & $<$0.001 & 99.0\% \\
gemini-2.5-flash & Extended & 7.380 & 7.189 & 7.577 & $<$0.001 & 98.3\% \\
gemini-2.5-pro & Extended & 7.694 & 7.482 & 7.906 & $<$0.001 & 98.9\% \\
gpt-oss-120b & Extended & 6.589 & 6.399 & 6.783 & $<$0.001 & 97.6\% \\
gpt-5 & Extended & 7.523 & 7.325 & 7.724 & $<$0.001 & 98.6\% \\
gpt-oss-20b & Extended & 5.438 & 5.259 & 5.616 & $<$0.001 & 95.9\% \\
glm-4.5 & Extended & 6.788 & 6.594 & 6.977 & $<$0.001 & 98.2\% \\
\bottomrule
\end{longtable}
\endgroup

\clearpage
\subsection{Parameter recovery and Parameter dispersion over fits}
\label{app:behav_recov}
\begin{figure}[htp]
    \centering
    \includegraphics[width=1\textwidth]{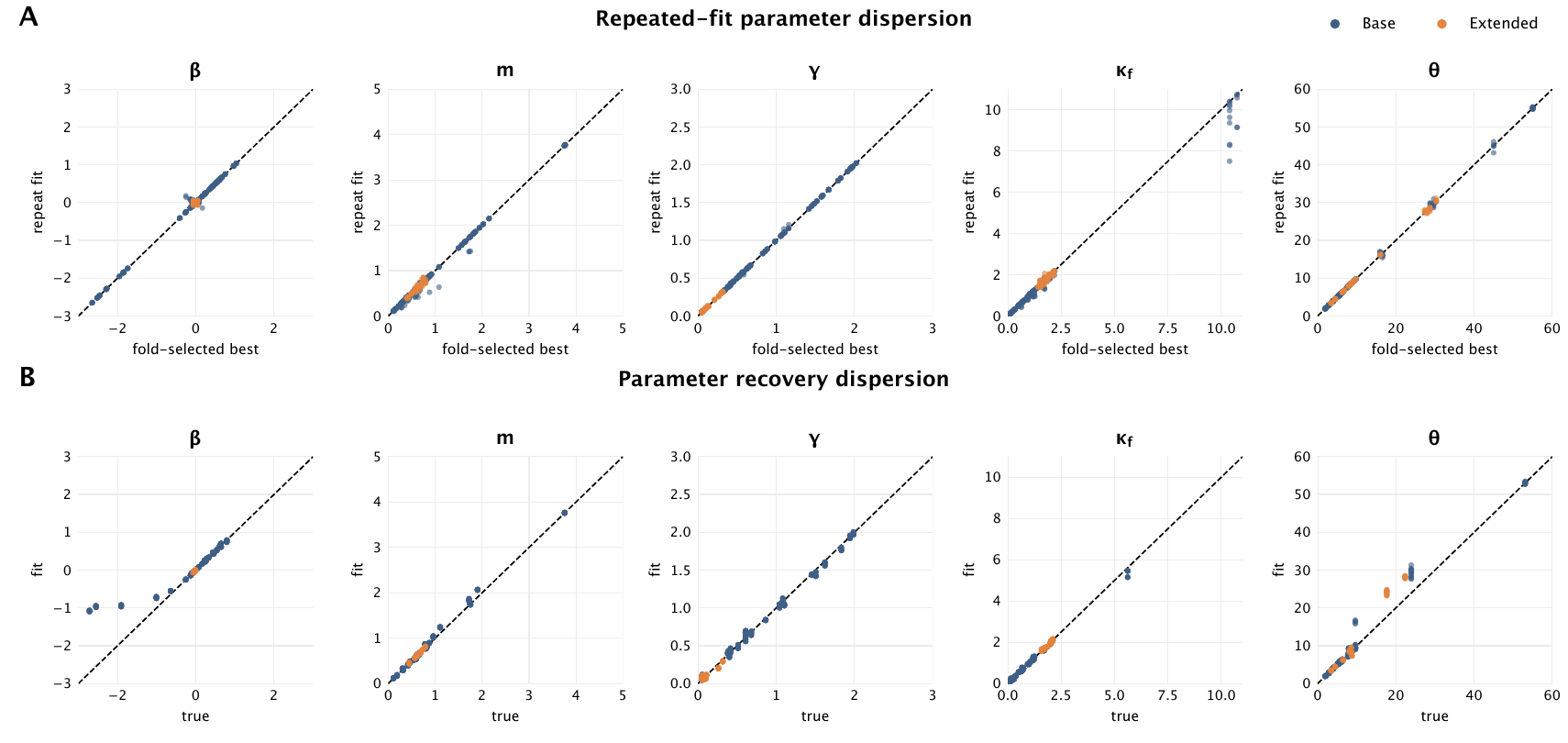}
    \caption{\textbf{A}, Fit quality and parameter dispersion across repeated fits. For each agent, the model was fit repeatedly on held-out data splits with multiple random initializations, retaining the best fit within each split by held-out log-likelihood. Each dot is one of the 15 repeated fits per agent (3 cross-validation folds $\times$ 5 initializations). \textbf{B}, Parameter recovery for the best model across agents. For each agent, synthetic datasets were generated from the selected fitted parameters. The model was then refit using the same fitting procedure; each dot is a recovered parameter from one of the 15 refits (3 folds $\times$ 5 initializations). }
    \label{fig:parameter_rec_disp}
\end{figure}

\clearpage
\subsection{Pairwise fit-quality heatmaps}
\label{app:behav_pairwise}
\begin{figure}[H]
    \centering
    \includegraphics[width=0.78\textwidth]{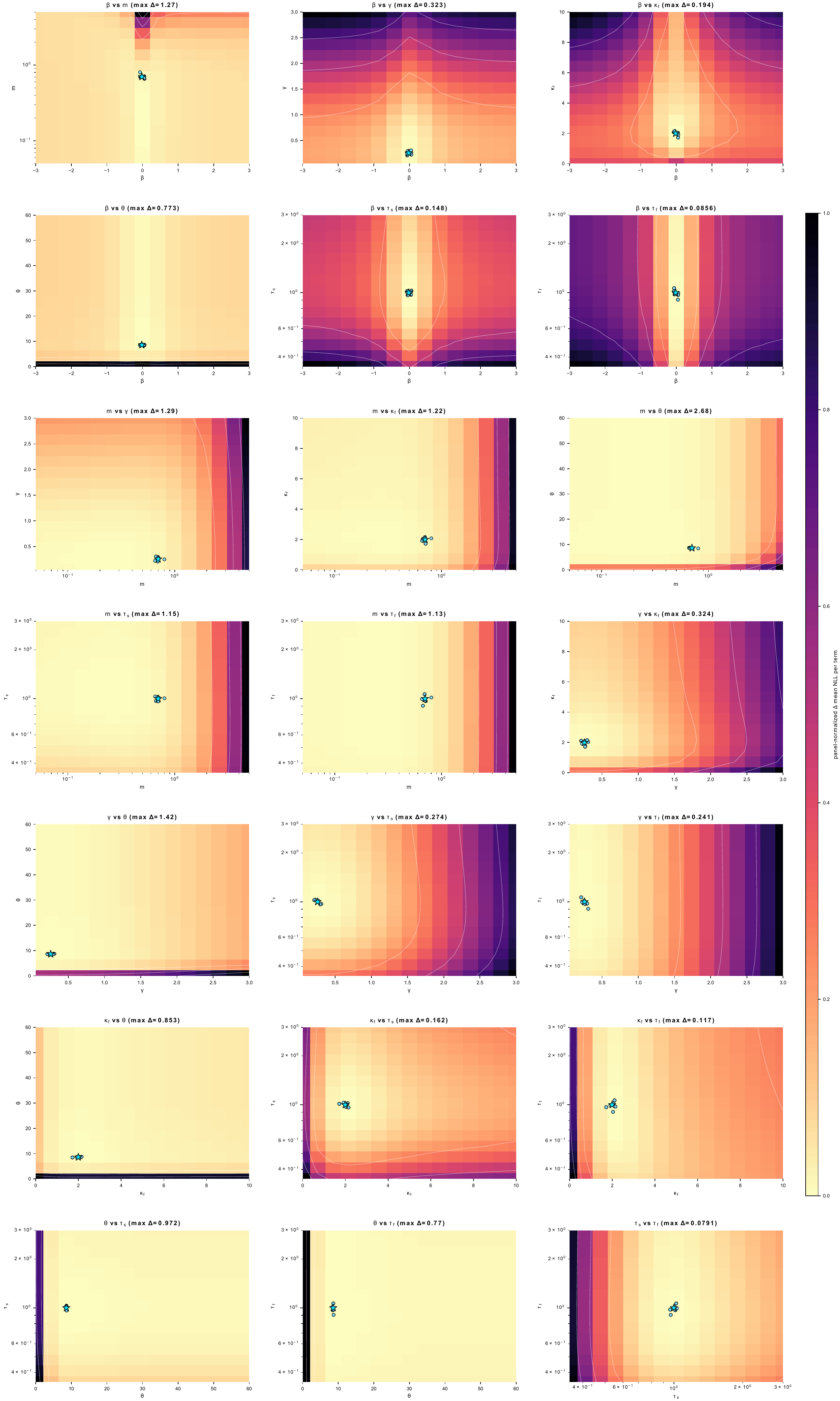}
    \caption{\textbf{Pairwise likelihood landscapes \textit{gpt-5-mini Extended}:} All parameters were fixed at their best fitted values except for one pair, which was varied over a two-dimensional grid while evaluating the negative log-likelihood of the empirical data. $\tau_s$ and $\tau_f$ are not model parameters: they rescale the fitted sampling and final-choice bias vectors, $\bm\omega_s^{(\tau)}\propto\bm\omega_s^{1/\tau_s}$ and $\bm\omega_f^{(\tau)}\propto\bm\omega_f^{1/\tau_f}$ (renormalized), so $\tau=1$ is the fitted bias, $\tau<1$ sharpens it and $\tau>1$ flattens it toward uniform (log-spaced grid, $\tau\in[0.35,3]$); for each repeated fit, $\tau$ is set by the inverse slope of its centered log-bias against the fitted one. Each dot is one of the 15 repeated fits (3 folds $\times$ 5 initializations); the star marks the fold-averaged fit.}
    \label{fig:pairwise_gpt}
\end{figure}
\clearpage
\begin{figure}[H]
    \centering
    \includegraphics[width=0.78\textwidth]{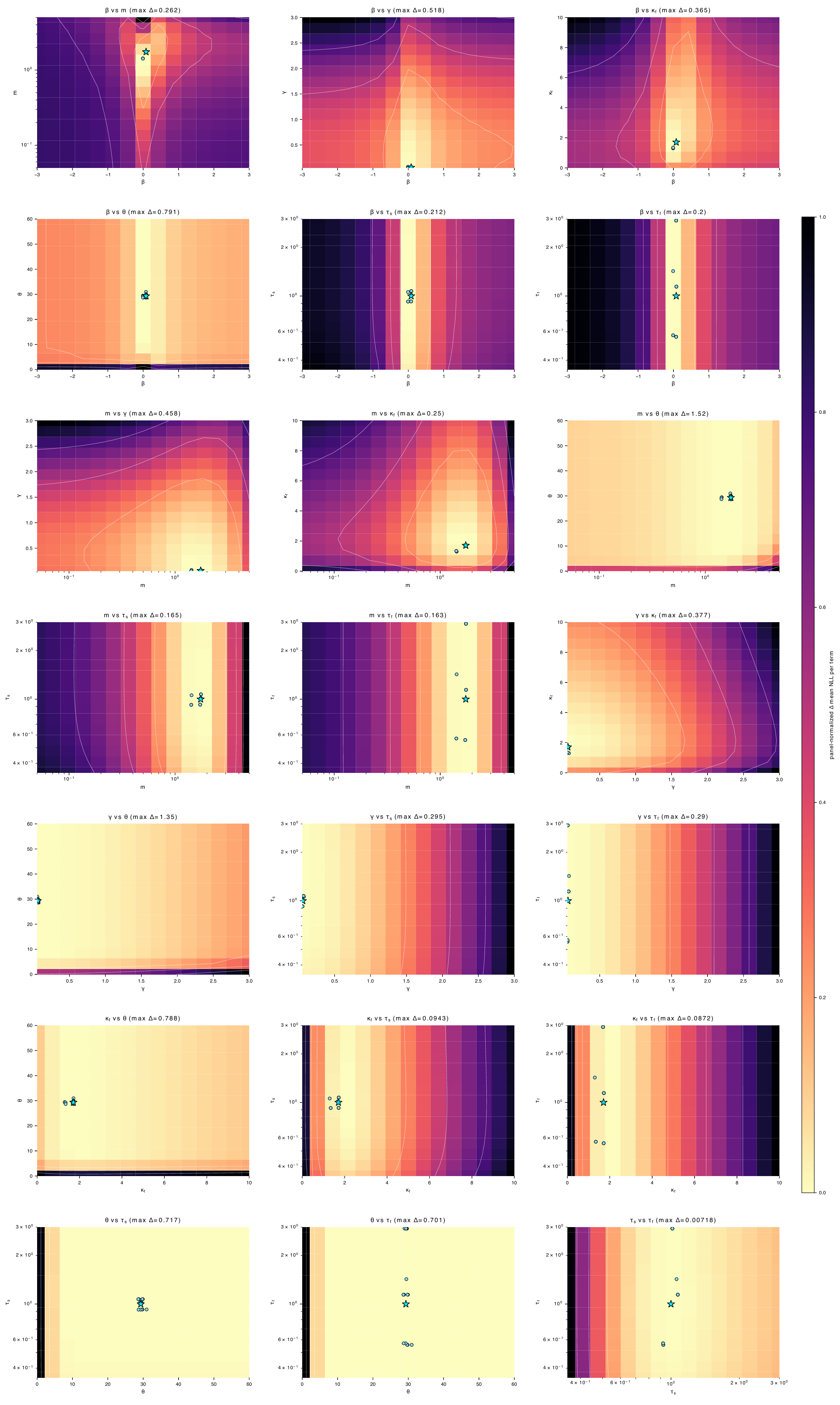}
    \caption{\textbf{Pairwise likelihood landscapes \textit{Humans}:} All parameters were fixed at their best fitted values except for one pair, which was varied over a two-dimensional grid while evaluating the negative log-likelihood of the empirical data. $\tau_s$ and $\tau_f$ are not model parameters: they rescale the fitted sampling and final-choice bias vectors, $\bm\omega_s^{(\tau)}\propto\bm\omega_s^{1/\tau_s}$ and $\bm\omega_f^{(\tau)}\propto\bm\omega_f^{1/\tau_f}$ (renormalized), so $\tau=1$ is the fitted bias, $\tau<1$ sharpens it and $\tau>1$ flattens it toward uniform (log-spaced grid, $\tau\in[0.35,3]$); for each repeated fit, $\tau$ is set by the inverse slope of its centered log-bias against the fitted one. Each dot is one of the 15 repeated fits (3 folds $\times$ 5 initializations); the star marks the fold-averaged fit.}
    \label{fig:pairwise_human}
\end{figure}
\clearpage
\begin{figure}[H]
    \centering
    \includegraphics[width=0.78\textwidth]{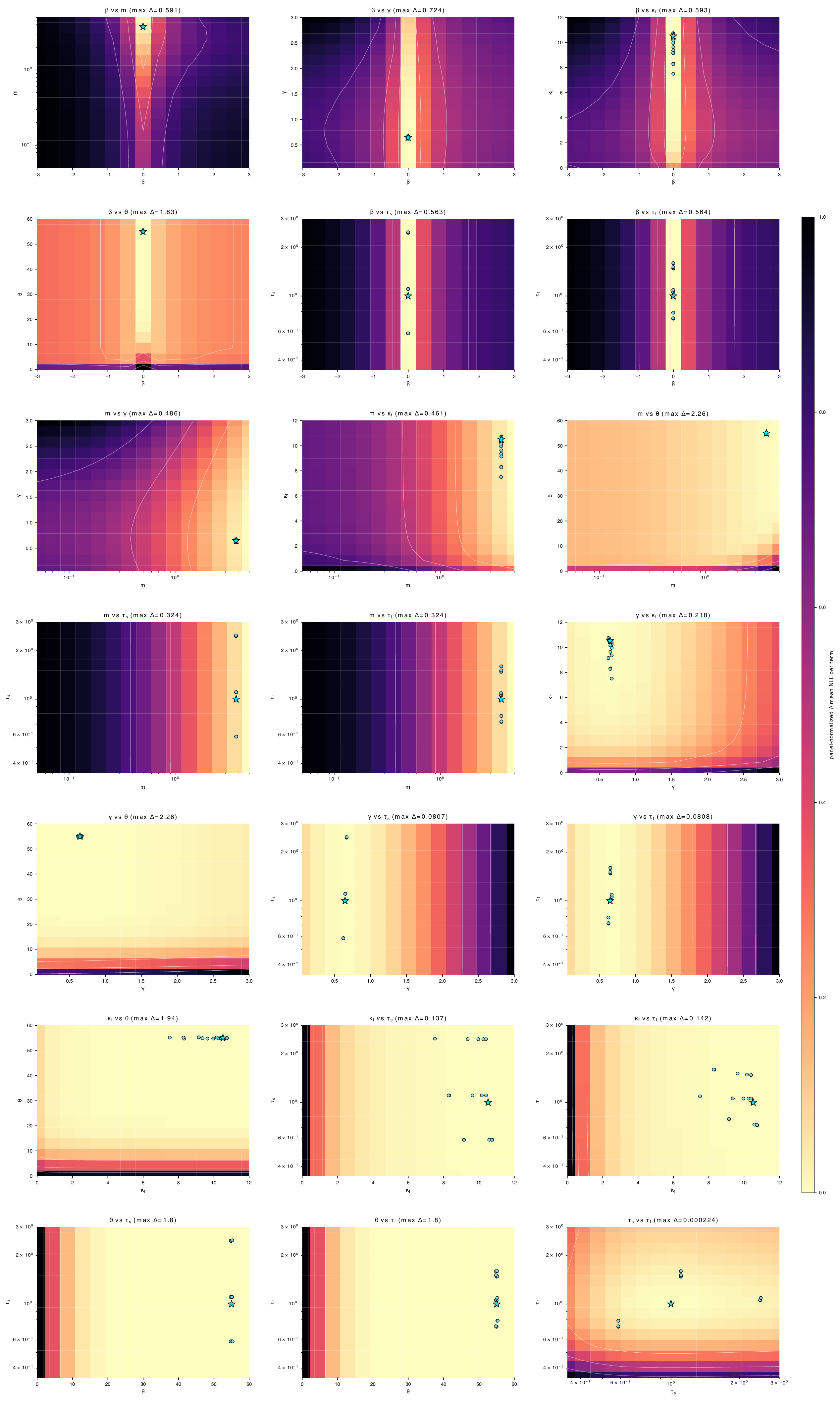}
    \caption{\textbf{Pairwise likelihood landscapes \textit{Optimal Agent (DP)}:} All parameters were fixed at their best fitted values except for one pair, which was varied over a two-dimensional grid while evaluating the negative log-likelihood of the empirical data. $\tau_s$ and $\tau_f$ are not model parameters: they rescale the fitted sampling and final-choice bias vectors, $\bm\omega_s^{(\tau)}\propto\bm\omega_s^{1/\tau_s}$ and $\bm\omega_f^{(\tau)}\propto\bm\omega_f^{1/\tau_f}$ (renormalized), so $\tau=1$ is the fitted bias, $\tau<1$ sharpens it and $\tau>1$ flattens it toward uniform (log-spaced grid, $\tau\in[0.35,3]$); for each repeated fit, $\tau$ is set by the inverse slope of its centered log-bias against the fitted one. Each dot is one of the 15 repeated fits (3 folds $\times$ 5 initializations); the star marks the fold-averaged fit.}
    \label{fig:pairwise_dp}
\end{figure}

\clearpage
\subsection{Model Fitting Parameters}
\label{appdx:model_parameter_table}
\begingroup
\tiny
\setlength{\tabcolsep}{3pt}
\renewcommand{\arraystretch}{0.95}
\begin{longtable}{@{} p{2.7cm} p{0.95cm} p{2.95cm} p{2.95cm} r r r r r @{}}
\caption{Fitted parameters of the gamma memory model. Values are best-by-fold averages from the selected fit.} \label{tab:memorybeta_fit_params} \\
\toprule
Model & Reasoning & $\omega_s$ & $\omega_f$ & $\beta$ & $m$ & $\gamma$ & $\kappa_f$ & $\theta_s$ \\
\midrule
\endfirsthead
\toprule
Model & Reasoning & $\omega_s$ & $\omega_f$ & $\beta$ & $m$ & $\gamma$ & $\kappa_f$ & $\theta_s$ \\
\midrule
\endhead
\midrule
\multicolumn{9}{r}{Continued on next page} \\
\midrule
\endfoot
\bottomrule
\endlastfoot
Optimal agent (DP) & -- & [0.252, 0.250, 0.246, 0.252] & [0.238, 0.220, 0.251, 0.292] & -0.004 & 3.763 & 0.646 & 10.513 & 55.004 \\
Humans & -- & [0.279, 0.268, 0.232, 0.221] & [0.257, 0.261, 0.239, 0.243] & 0.081 & 1.742 & 0.059 & 1.707 & 29.322 \\
gpt-5-mini & Base & [0.594, 0.285, 0.098, 0.023] & [0.467, 0.317, 0.160, 0.056] & -0.101 & 0.602 & 0.415 & 1.230 & 5.676 \\
gpt-5-mini & Extended & [0.359, 0.281, 0.232, 0.129] & [0.364, 0.294, 0.208, 0.135] & -0.019 & 0.701 & 0.257 & 1.992 & 8.590 \\
grok-3-mini & Base & [0.289, 0.329, 0.207, 0.176] & [0.392, 0.273, 0.192, 0.143] & -0.012 & 0.665 & 0.050 & 1.709 & 12.040 \\
grok-3-mini & Extended & [0.313, 0.309, 0.216, 0.163] & [0.379, 0.304, 0.185, 0.132] & 0.013 & 0.675 & 0.050 & 1.941 & 9.119 \\
gemini-2.5-flash & Base & [0.741, 0.160, 0.070, 0.029] & [0.646, 0.179, 0.101, 0.074] & 0.255 & 0.605 & 1.046 & 0.699 & 9.034 \\
gemini-2.5-flash & Extended & [0.399, 0.273, 0.195, 0.133] & [0.384, 0.286, 0.189, 0.140] & -0.056 & 0.622 & 0.093 & 1.740 & 8.166 \\
gemini-2.5-pro & Base & [0.613, 0.220, 0.125, 0.043] & [0.493, 0.213, 0.154, 0.141] & 0.071 & 0.512 & 0.595 & 0.539 & 30.369 \\
gemini-2.5-pro & Extended & [0.508, 0.237, 0.162, 0.093] & [0.476, 0.214, 0.178, 0.132] & -0.037 & 0.792 & 0.106 & 1.613 & 29.179 \\
gpt-oss-120b & Base & [0.427, 0.181, 0.363, 0.030] & [0.468, 0.217, 0.278, 0.037] & -0.089 & 0.420 & 0.513 & 1.184 & 2.879 \\
gpt-oss-120b & Extended & [0.320, 0.275, 0.199, 0.205] & [0.425, 0.251, 0.164, 0.160] & -0.013 & 0.599 & 0.050 & 1.610 & 4.486 \\
gpt-5 & Base & [0.518, 0.275, 0.132, 0.075] & [0.428, 0.284, 0.158, 0.130] & -0.037 & 0.148 & 1.452 & 0.674 & 5.052 \\
gpt-5 & Extended & [0.295, 0.282, 0.239, 0.183] & [0.371, 0.267, 0.214, 0.148] & -0.005 & 0.625 & 0.051 & 2.055 & 23.948 \\
gpt-oss-20b & Base & [0.358, 0.284, 0.217, 0.141] & [0.344, 0.277, 0.216, 0.163] & -2.337 & 2.039 & 0.552 & 0.248 & 2.898 \\
gpt-oss-20b & Extended & [0.338, 0.296, 0.207, 0.159] & [0.385, 0.268, 0.190, 0.156] & -0.047 & 0.446 & 0.316 & 1.601 & 3.475 \\
claude-sonnet-4 & Base & [0.399, 0.265, 0.184, 0.152] & [0.508, 0.217, 0.140, 0.135] & -0.032 & 0.438 & 0.050 & 2.038 & 7.965 \\
glm-4.5 & Base & [0.266, 0.418, 0.233, 0.083] & [0.416, 0.318, 0.175, 0.091] & 0.224 & 0.309 & 0.856 & 0.660 & 3.884 \\
glm-4.5 & Extended & [0.344, 0.276, 0.244, 0.137] & [0.358, 0.262, 0.262, 0.118] & -0.020 & 0.569 & 0.078 & 1.878 & 6.503 \\
DeepSeek-R1-0528-Qwen3-8B & Base & [0.244, 0.324, 0.302, 0.130] & [0.230, 0.314, 0.296, 0.159] & 0.014 & 0.418 & 0.405 & 1.050 & 5.307 \\
gpt4.1-mini & Base & [0.596, 0.263, 0.072, 0.069] & [0.536, 0.258, 0.091, 0.114] & -0.068 & 0.659 & 0.578 & 1.170 & 5.163 \\
claude-haiku-3.5 & Base & [0.175, 0.370, 0.359, 0.096] & [0.248, 0.400, 0.242, 0.110] & -0.133 & 0.305 & 0.387 & 0.936 & 5.599 \\
Qwen3 235B Instruct & Base & [0.451, 0.334, 0.141, 0.074] & [0.416, 0.320, 0.160, 0.104] & 0.649 & 0.113 & 0.050 & 1.243 & 7.934 \\
gpt4o-mini & Base & [0.613, 0.284, 0.080, 0.023] & [0.418, 0.393, 0.126, 0.063] & -0.254 & 0.961 & 1.107 & 0.645 & 8.667 \\
Qwen 14B Instruct & Base & [0.469, 0.257, 0.140, 0.134] & [0.382, 0.229, 0.133, 0.256] & 0.464 & 0.318 & 0.050 & 1.001 & 8.791 \\
Llama-4-Scout-Instruct & Base & [0.621, 0.165, 0.115, 0.100] & [0.517, 0.223, 0.118, 0.142] & 0.435 & 0.803 & 1.842 & 0.714 & 5.389 \\
Qwen 7B Instruct & Base & [0.360, 0.297, 0.209, 0.134] & [0.522, 0.253, 0.118, 0.106] & 0.439 & 0.875 & 0.050 & 0.509 & 3.713 \\
Llama-3.1-405b-Instruct & Base & [0.481, 0.303, 0.139, 0.077] & [0.480, 0.290, 0.119, 0.111] & -0.252 & 0.661 & 0.064 & 0.446 & 7.783 \\
Llama-4-Maverick-Instruct & Base & [0.520, 0.215, 0.179, 0.085] & [0.422, 0.249, 0.203, 0.126] & 0.640 & 0.637 & 1.956 & 0.186 & 5.954 \\
Llama 3.3 70B Instruct & Base & [0.387, 0.372, 0.148, 0.092] & [0.359, 0.372, 0.139, 0.131] & -0.109 & 0.897 & 1.054 & 0.069 & 4.394 \\
Gemma 2B Instruct & Base & [0.354, 0.207, 0.327, 0.112] & [0.277, 0.234, 0.368, 0.121] & 0.527 & 0.557 & 0.050 & 0.192 & 3.966 \\
Centaur 70B & Base & [0.825, 0.156, 0.015, 0.004] & [0.729, 0.263, 0.004, 0.004] & 0.062 & 0.695 & 1.640 & 0.066 & 6.508 \\
Centaur 8B & Base & [0.366, 0.249, 0.281, 0.103] & [0.294, 0.175, 0.349, 0.183] & 0.598 & 0.621 & 1.987 & 0.006 & 6.278 \\
Qwen3 32B & Base & [0.517, 0.289, 0.148, 0.046] & [0.419, 0.323, 0.196, 0.062] & -0.097 & 0.873 & 1.564 & 0.021 & 6.214 \\
Apertus 70B Instruct & Base & [0.304, 0.298, 0.189, 0.209] & [0.219, 0.305, 0.198, 0.278] & -1.849 & 1.615 & 0.050 & 0.014 & 4.054 \\
Apertus 8B Instruct & Base & [0.384, 0.268, 0.180, 0.168] & [0.226, 0.318, 0.150, 0.306] & -2.548 & 1.752 & 0.381 & 0.134 & 2.194 \\
Qwen 3B Instruct & Base & [0.275, 0.244, 0.280, 0.201] & [0.240, 0.166, 0.295, 0.299] & 0.516 & 0.705 & 0.050 & 0.005 & 1.951 \\
Llama-3.2-3B-Instruct & Base & [0.363, 0.224, 0.208, 0.205] & [0.162, 0.170, 0.321, 0.347] & 0.539 & 0.823 & 0.050 & 0.314 & 3.139 \\
\end{longtable}
\endgroup

\begingroup
\small
\begin{longtable}{llrrr}
\caption{Negative log-likelihoods of the mechanistic model repeated fold fits. Values are averages over the 15 repeated fold fits per model-condition.} \label{tab:memorybeta_nll_repeated_folds} \\
\toprule
Model & Reasoning & Train NLL & Test NLL & NLL \\
\midrule
\endfirsthead
\caption[]{Negative log-likelihoods of the mechanistic model repeated fold fits.} \\
\toprule
Model & Reasoning & Train NLL & Test NLL & NLL \\
\midrule
\endhead
\midrule
\multicolumn{5}{r}{Continued on next page} \\
\midrule
\endfoot
\bottomrule
\endlastfoot
Optimal agent (DP) & -- & 0.303 & 0.303 & 0.303 \\
Humans & -- & 0.657 & 0.657 & 0.657 \\
gpt-5-mini & Base & 0.613 & 0.615 & 0.615 \\
gpt-5-mini & Extended & 0.653 & 0.654 & 0.654 \\
grok-3-mini & Base & 0.664 & 0.666 & 0.666 \\
grok-3-mini & Extended & 0.710 & 0.712 & 0.712 \\
gemini-2.5-flash & Base & 0.539 & 0.541 & 0.541 \\
gemini-2.5-flash & Extended & 0.676 & 0.679 & 0.679 \\
gemini-2.5-pro & Base & 0.628 & 0.627 & 0.627 \\
gemini-2.5-pro & Extended & 0.610 & 0.612 & 0.612 \\
gpt-oss-120b & Base & 0.945 & 0.946 & 0.946 \\
gpt-oss-120b & Extended & 0.796 & 0.797 & 0.797 \\
gpt-5 & Base & 0.712 & 0.712 & 0.712 \\
gpt-5 & Extended & 0.650 & 0.651 & 0.651 \\
gpt-oss-20b & Base & 1.055 & 1.059 & 1.059 \\
gpt-oss-20b & Extended & 0.901 & 0.904 & 0.904 \\
claude-sonnet-4 & Base & 0.668 & 0.669 & 0.669 \\
glm-4.5 & Base & 0.912 & 0.913 & 0.913 \\
glm-4.5 & Extended & 0.922 & 0.924 & 0.924 \\
DeepSeek-R1-0528-Qwen3-8B & Base & 0.846 & 0.847 & 0.847 \\
gpt4.1-mini & Base & 0.661 & 0.663 & 0.663 \\
claude-haiku-3.5 & Base & 0.750 & 0.753 & 0.753 \\
Qwen3 235B Instruct & Base & 0.693 & 0.694 & 0.694 \\
gpt4o-mini & Base & 0.543 & 0.546 & 0.546 \\
Qwen 14B Instruct & Base & 0.725 & 0.727 & 0.727 \\
Llama-4-Scout-Instruct & Base & 0.735 & 0.737 & 0.737 \\
Qwen 7B Instruct & Base & 0.887 & 0.888 & 0.888 \\
Llama-3.1-405b-Instruct & Base & 0.733 & 0.734 & 0.734 \\
Llama-4-Maverick-Instruct & Base & 0.750 & 0.753 & 0.753 \\
Llama 3.3 70B Instruct & Base & 0.799 & 0.798 & 0.798 \\
Gemma 2B Instruct & Base & 0.875 & 0.876 & 0.876 \\
Centaur 70B & Base & 0.493 & 0.491 & 0.491 \\
Centaur 8B & Base & 0.715 & 0.717 & 0.717 \\
Qwen3 32B & Base & 0.645 & 0.646 & 0.646 \\
Apertus 70B Instruct & Base & 0.890 & 0.892 & 0.892 \\
Apertus 8B Instruct & Base & 1.137 & 1.141 & 1.141 \\
Qwen 3B Instruct & Base & 1.297 & 1.299 & 1.299 \\
Llama-3.2-3B-Instruct & Base & 1.131 & 1.132 & 1.132 \\
\end{longtable}
\endgroup

\end{document}